\documentclass{esub2acm}
\usepackage{latexsym}
\usepackage{exscale}

\newtheorem{theorem}{Theorem}[section]
\newtheorem{corollary}[theorem]{Corollary}
\newtheorem{proposition}[theorem]{Proposition}
\newtheorem{lemma}[theorem]{Lemma}

\newtheorem{definition}[theorem]{Definition}
\newtheorem{remarkTh}[theorem]{Remark}
\newenvironment{remark}[0]{\begin{remarkTh}\rm}{\lqed\end{remarkTh}}
\newtheorem{exampleTh}[theorem]{Example}
\newenvironment{example}[0]{\begin{exampleTh}\rm}{\lqed\end{exampleTh}}

\newcommand{\lqed}{\hspace*{\fill}\mbox{$ \Box $}\vspace{0.15in}}
\newcommand{\text}[1]{\mbox{#1}}
\newcommand{\Not}{{\it not}\,}
\newcommand{\sneg}{{\mbox{{--\vspace{.07in}}}\!}}   

\newcounter{itemlistDepth}

{\addtocounter{itemlistDepth}{1}%
\ifnum2<\value{itemlistDepth}\errmessage{items nesting too deep!}\fi%
\ifnum1=\value{itemlistDepth}%
\begin{list}{\(\bullet\)\hfill}{
\parskip 3pt plus 1pt
\labelwidth 10pt
\labelsep 0pt 
\leftmargin 1.5em \advance\leftmargin\labelwidth
\listparindent \parindent
\topsep 0pt
\parsep 3pt plus 1pt
\itemsep 0pt  
\partopsep 0pt
\itemindent 0pt
\rightmargin 0pt
}\fi%
\ifnum2=\value{itemlistDepth}%
\begin{list}{-\hfill}{
\parskip 3pt plus 1pt
\labelwidth 10pt
\labelsep 0pt
\leftmargin 1.5em \advance\leftmargin\labelwidth
\listparindent \parindent
\topsep 0pt
\parsep 3pt plus 1pt
\itemsep 0pt
\partopsep 0pt
\itemindent 0pt
\rightmargin 0pt
}\fi%
}%
{\end{list}\addtocounter{itemlistDepth}{-1}}

\catcode`\<=\active
\def<{\(}

\catcode`\>=\active
\def>{\)}

\newcommand{\mathsym}[1]{\relax\ifmmode#1\else
        \errmessage{Mathematical symbol outside math mode}\fi}
\newcommand{\mathid}[1]{\mathsym{{\it #1}}}

\newcommand{\lt}{\mathsym{\mathchar 12604\relax}} 

\newcommand{\union}{\cup}
\newcommand{\intersect}{\cap}
\newcommand{\setCond}[2]{\{#1\,:\,#2\}}
\newcommand{\setCondB}[2]{\bigl\{#1\,:\,#2\bigr\}}
\newcommand{\set}[1]{\{#1\}}
\newcommand{\setB}[1]{\bigl\{#1\bigr\}}

\newcommand{\natNum}{\mathsym{{\rm I\kern-0.17em N}}}
\renewcommand{\i}{\mathsym{i}}

\ifx\k\undefined
        \newcommand{\k}{\mathsym{k}}
\else
        \renewcommand{\k}{\mathsym{k}}
\fi
\newcommand{\n}{\mathsym{n}}
\newcommand{\m}{\mathsym{m}}

\newcommand{\defEq}{:=}
\newcommand{\until}{, \ldots,}
\newcommand{\metaThen}{\mathsym{\;\Longrightarrow\;}}

\newcommand{\struct}[1]{(#1)}
\newcommand{\defIff}{\;:\Longleftrightarrow\;}
\newcommand{\metaIf}{\;\Longleftarrow\;}

\newcommand{\parenB}[1]{\bigl(#1\bigr)}
\newcommand{\parenC}[1]{\Bigl(#1\Bigr)}
\newcommand{\parenD}[1]{\biggl(#1\biggr)}

\newcommand{\F}{\mathsym{F}}        
\newcommand{\G}{\mathsym{G}}
\newcommand{\E}{\mathsym{E}}
\newcommand{\p}{\mathsym{p}}        
\newcommand{\q}{\mathsym{q}}        
\ifx\r\undefined
        \newcommand{\r}{\mathsym{r}}        
\else
        \renewcommand{\r}{\mathsym{r}}      
\fi
\newcommand{\lorUntil}{\lor\cdots\lor}
\newcommand{\landUntil}{\land\cdots\land}

\newcommand{\landMulti}[2]{\bigwedge_{#1}^{#2}}
\newcommand{\lneg}{\mathsym{\lnot}}
\newcommand{\lif}{\mathsym{\gets}}
\newcommand{\liff}{\mathsym{\leftrightarrow}}
\newcommand{\lthen}{\mathsym{\to}}

\newcommand{\ruleEnd}{.}

\newcommand{\false}{{\it false}}
\newcommand{\true}{{\it true}}

\newcommand{\A}{\mathsym{A}}

\renewcommand{\P}{\mathsym{P}}

\newcommand{\At}{At_{\mathcal L}}
\newcommand{\Lang}{{\mathcal L}}
\newcommand{\CnNot}{{\it Cn}_{\it not}}
\newcommand{\Cn}{{\it Cn}}

\newcommand{\LNot}{\Lang_{\it not}}

\newcommand{\objI}{I_{\it obj}} 
\newcommand{\redI}{I}      
\newcommand{\Obj}{{\mathcal OBJ}} 
\newcommand{\objSet}{{\mathcal O}} 
\newcommand{\fullI}{{\mathcal I}}   

\newcommand{\defI}{I_{\it not}} 
\newcommand{\Def}{{\mathcal NOT}} 
\newcommand{\defSet}{{\mathcal N}} 
\newcommand{\DefToObj}[1]{\Omega_{#1}} 
\newcommand{\ObjToDef}[1]{\Pi_{#1}} 
\newcommand{\DefToDef}[1]{\Theta_{#1}} 
\newcommand{\DefFix}{{\mathcal N}^\diamond} 

\newcommand{\T}{T}            
\newcommand{\0}{\mbox{false}}            
\newcommand{\1}{\mbox{true}}            

\newcommand{\w}{\mathsym{w}}    
\newcommand{\K}{\mathsym{{\mathcal K}}}     
\newcommand{\W}{\mathsym{W}}    
\newcommand{\R}{\mathsym{R}}    
\newcommand{\V}{\mathsym{V}}    

\newcommand{\Car}{\mathid{car}}
\newcommand{\Broken}{\mathid{broken}}
\newcommand{\Runs}{\mathid{runs}}
\newcommand{\Fixed}{\mathid{fixed}}

\newcommand{\VisitEurope}{\mathid{visit\_europe}}
\newcommand{\VisitAustralia}{\mathid{visit\_australia}}
\newcommand{\Happy}{\mathid{happy}}
\newcommand{\Bankrupt}{\mathid{bankrupt}}
\newcommand{\Prudent}{\mathid{prudent}}
\newcommand{\Disappointed}{\mathid{disappointed}}

\newcommand{\Drink}{\mathid{drink}}
\newcommand{\Drive}{\mathid{drive}}

\newcommand{\BB}{{{\mathcal B}}}
\newcommand{\mh}[1]{\ensuremath{#1}}       

\newcounter{savedSection}
\newcounter{savedTheorem}

\begin{document}

\title{Super Logic Programs}
  
\author{Stefan Brass\\
        University of Pittsburgh,\\
        J\"{u}rgen Dix\\
        The University of Manchester
        \and
        Teodor C.~Przymusinski\\
        University of California, Riverside}


\begin{abstract}
  The Autoepistemic Logic of Knowledge and Belief (AELB) is a powerful
  nonmonotonic formalism introduced by Teodor Przymusinski in 1994.  In
  this paper, we specialize it to a class of theories called ``super
  logic programs''.  We argue that these programs form a natural
  generalization of standard logic programs.  In particular, they
  allow disjunctions and default negation of arbitrary positive
  objective formulas.
  
  Our main results are two new and important characterizations of the
  static semantics of these programs, one syntactic, and one
  model-theoretic.  The syntactic fixed point characterization is much
  simpler than the fixed point construction of the static semantics
  for arbitrary AELB theories.  The model-theoretic characterization
  via Kripke models allows one to construct finite representations of
  the inherently infinite static expansions.
  
  Both characterizations can be used as the basis of algorithms for
  query answering under the static semantics.  We describe 
   a \emph{query-answering interpreter} for super
  programs which we developed based on the model-theoretic
  characterization and which is available on the web.
\end{abstract}


[Logic and constraint programming, Modal logic]
[Logic Programming, Nonmonotonic reasoning and belief revision]


\begin{bottomstuff}
  This article is a full version of some of the results reported in an
  extended abstract which appeared in the Proceedings of the Fifth
  International Conference on Principles of Knowledge Representation
  and Reasoning (KR'96), L.~C.~Aiello,
J.~Doyle and S.~C.~Shapiro (editors), 
1996, pp. 529--541.

\begin{authinfo}
\name{Stefan Brass}
\affiliation{University of Giessen, Inst.~f.~Informatik}
\address{Arndtstr.~2, 35392~Giessen, Germany, EMAIL: brass@acm.org}
\name{J\"{u}rgen Dix}
\affiliation{The University of Manchester, Dept.~of Computer Science}
\address{Oxford Road, Manchester M13 9PL, UK, EMAIL: dix@cs.man.ac.uk}
\name{Teodor C.~Przymusinski}
\affiliation{Dept.~of Computer Science, University of California, Riverside}
\address{CA 92521, USA, EMAIL: teodor@cs.ucr.edu}
\end{authinfo}
\end{bottomstuff}

\markboth{S.~Brass and J.~Dix and T.~C.~Przymusinski}{Super Logic Programs}
\maketitle


\section{Introduction}

The relationship between logic programs and non-monotonic knowledge
representation formalisms can be briefly summarized as follows. Any
non-monotonic formalism for knowledge representation has to contain
some form of \emph{default negation}, whether it is defined as an
explicit negation operator or is implicitly present in the form of
beliefs, disbeliefs or defaults. Since the main difference between
logic programs and classical (monotonic) logic is the presence of
default negation, logic programs can be viewed, in this sense, as the
simplest non-monotonic extension of classical monotonic logic.


However, standard logic programs suffer from some important
limitations. Most importantly, they are unable to handle disjunctive
information%
\footnote{The stable model semantics permits multiple models,
which is something similar.
But it has to be used with caution,
since it can easily become inconsistent.
Depending on the application,
disjunctions can often be more natural.}.
Yet, in natural discourse as well as in various
programming applications we often use {\em disjunctive statements}. One
particular example of such a situation is {\em reasoning by cases.} Other
obvious examples include:
\begin{enumerate}
\item \emph{Approximate information:} for instance, an age
  ``around~30'' can be~28, 29, 30, 31, or~32;
\item \emph{Legal rules:} the judge always has some freedom for his
  decision, otherwise he/she would not be needed; so laws cannot have
  unique models;
\item \emph{Diagnosis:} only at the end of a fault diagnosis we may
  know exactly which part of some machine was faulty but as long as we
  are searching, different possibilities exist;
\item \emph{Biological inheritance:} ~if the parents have blood
  groups~A and~0, the child must also have one of these two blood
  groups (example from~\cite{Lip79});
\item \emph{Natural language understanding:} here there are many
  possibilities for ambiguity and they are represented most naturally
  by multiple intended models;
\item \emph{Reasoning about concurrent processes:} since we do not
  know the exact sequence in which certain operations are performed,
  again multiple models come into play;

\item \emph{Conflicts in multiple inheritance:} if we want to keep
  as much information as possible, we should assume disjunction
  of the inherited values, see~\cite{BL93Defaults}.
\end{enumerate}
Disjunctive logic programming was used e.g.~in the following projects
for real applications:
\begin{enumerate}
\item In the DisLoP project~\cite{AraDixNie97comp}, a logical
  description of a system together with a particular observation is
  transformed into a disjunctive program so that an interpreter for
  disjunctive logic programs can compute the minimal diagnosis
  \cite{Baumgartner:etal:TableauxDiagnosis:IJCAI:97}.
  \item The view update problem in deductive and relational databases
  can also be transformed into a problem of computing minimal models
  of disjunctive programs
  \cite{Aravindan:Baumgartner:ViewDeletion:ILPS:97}.
\item Another important application of disjunctive techniques in the
  database area is to glue together different heterogeneous databases
  to provide a single unified view to the user.  With the ever
  expanding world wide web technology, millions and millions of data
  in thousands of different formats are thrown at a user who clearly
  needs some tools to put together information of interest to him.
  In \cite{SchNeu97,Tho98} it is shown how disjunctive logic programs
 can be used  as mediators and thus as an important tool to create a
 powerful query language for accessing the web.
\item In~\cite{ST96,Sto98} disjunctive logic programming was used for
  analyzing rule sets for calculating banking fees.
  A credit institution sells stocks and shares to its customers
  and charges their accounts. The fee depends on the value of the transaction,
  the customer type and various other parameters. All this is formulated
  in a set of rules in natural language. The problem is to translate
  this rule set into a formal language and then to analyze its various
  properties. For example does the rule set allow us to calculate
  a fee for each business deal? Is that fee unique? A straightforward
  translation into propositional logic and then using a theorem prover
  did not work. However, it turned out that using disjunctive logic
  programs solved the problem~\cite{ST96,Sto98}. 
\end{enumerate}

Formalisms promoting disjunctive reasoning are more {\em expressive} and
{\em natural to use} since they permit direct translation of disjunctive
statements from natural language and from informal specifications.
Consequently, considerable interest and research effort{\footnote{It
    suffices just to mention several recent workshops on
    \emph{Extensions of Logic Programming} specifically devoted to
    this subject
    (\cite{DPP:NMELP95,DPP:NMELP97,DycHerSch96,DixPerPrz98}).}} has
been recently given to the problem of finding a suitable extension of
the \emph{logic programming paradigm} beyond the class of normal logic
programs that would ensure a proper treatment of \emph{disjunctive
  information}.  However, the problem of finding a suitable semantics
for disjunctive programs and deductive databases proved to be far more
complex than it is in the case of normal, non-disjunctive
programs{\footnote{The book by Minker et. al.  \cite{LobMinRaj92}
    provides a detailed account of the extensive research effort in
    this area. See also
    \cite{Dix93semantics,Min93an,P:disjsem,P:static,Min96}.}}.

We believe that in order to demonstrate that a class of programs can be
justifiably called an \emph{extension of logic programs} one should be
able to argue that:
\begin{enumerate}
\item the proposed \emph{syntax} of such programs resembles the syntax
  of logic programs but it applies to a significantly broader class of
  programs, which includes the class of disjunctive logic programs as
  well as the class of logic programs with strong (or ``classical")
  negation \cite{GL:classic,APP:neg-ful};
\item the proposed \emph{semantics} of such programs constitutes an
  intuitively natural extension of one (or more) well-established
  semantics of normal logic programs;
\item there exists a reasonably simple procedural mechanism
  allowing, at least in principle, to compute the
  semantics{\footnote{Observe that while such a mechanism cannot --
      even in principle -- be efficient, due to the inherent
      NP-completeness of the problem of computing answers just
      to positive disjunctive programs, it can be efficient when
      restricted to specific subclasses of programs and queries
      and it can allow efficient approximation methods for
      broader classes of programs.}};
\item the proposed class of programs and their semantics is a
  special case of a more general non-monotonic formalism which would
  clearly link it to other well-established non-monotonic formalisms.
\end{enumerate}

In this paper we propose a specific class of such extended logic
programs which will be (immodestly) called \emph{super logic programs}
or just \emph{super-programs}. We will argue that the class of
super-programs satisfies all of the above conditions, and, in
addition, is sufficiently flexible to allow various
application-dependent extensions and modifications.
It includes all (monotonic) \emph{propositional theories}, all
\emph{disjunctive logic programs} and all \emph{extended logic programs}
with strong negation.
We also provide a
description of an implementation of a \emph{query-answering
  interpreter} for the class of super-programs which is available on the
WWW\footnote{See {\tt http://www.informatik.uni-giessen.de/staff/brass/slp/}.
If this URL should be unavailable,
try {\tt http://www2.sis.pitt.edu/{\char126}sbrass/slp/}.}.


The class of super logic programs is closely related to other non-monotonic formalisms.  In fact, super logic programs constitute a special case  of a yet more
expressive non-monotonic formalism called the \emph{Autoepistemic Logic
  of Knowledge and Beliefs}, \emph{AELB}, introduced earlier in
\cite{P:ael-min,P:ael-minf}.  The logic \emph{AELB} isomorphically includes the
well-known non-monotonic formalisms of Moore's \emph{Autoepistemic
  Logic} and McCarthy's \emph{Circumscription}. Through this embedding, the
semantics of super programs is clearly  linked to other
well-established non-monotonic formalisms.

At the same time, as we demonstrate in this paper, in spite of their
increased expressiveness, super logic programs still admit natural and
simple {\em query answering mechanisms} which can be easily implemented
and tested. We discuss one such existing implementation in this paper.
Needless to say, the problem of finding suitable inference mechanisms,
capable to model human common-sense reasoning, is one of the major
research and implementation problems in Artificial Intelligence.

The paper is organized as follows. In Section~\ref{sec:aeb} we recall
the definition and basic properties of \emph{non-monotonic knowledge bases},
and we introduce the class of super logic programs as a special subclass
  of them. We also establish
basic properties of super programs.  In the following
Sections~\ref{sec:fixed} and \ref{sec:model} we describe two 
characterizations of the semantics of super-programs, one of which is
syntactic and the other model-theoretic.  Due to the restricted nature
of super programs, these characterization are significantly simpler
than those applicable to arbitrary non-monotonic knowledge bases.  In
Section~\ref{sec:computation} we  describe our implementation
of a \emph{query-answering interpreter} for super-programs which is
based on the previously established model-theoretic characterization
of their semantics.  Section~\ref{sec:relwork} mentions related work.
We conclude with some final remarks in Section~\ref{sec:conclusions}.
For the sake of clarity, most proofs are contained in the Appendix.

Throughout the paper, we restrict our attention to
\emph{propositional} programs.  Since we can always consider a
propositional instantiation of the program, this does not limit the
generality of the obtained results from a semantic standpoint.
Of course, many practical applications need rules with variables.
The current  version of our interpreter for super-logic programs
is already supporting variables
that satisfy the allowedness/range-restriction condition
(at least one occurrence of each variable must be in a positive body literal).
This is done via an intelligent grounding mechanism,
see Section~\ref{sec:computation}.

\section{Super Logic Programs}
\label{sec:aeb}

We first define the notion of a \emph{non-monotonic knowledge base}.
Super logic programs are special knowledge bases.

The language is based on the \emph{Autoepistemic Logic of Knowledge
  and Beliefs}, \emph{AELB}, introduced in
\cite{P:ael-min,P:ael-minf}.  However, in this paper we use the
default negation operator~<\Not> instead of the belief operator~<\BB>.
Actually, <\Not\F> can be seen as an abbreviation for <\BB(\lneg\F)>, so
this is only a slight syntactic variant%
\footnote{$\Not\F$ was introduced as an abbreviation for <\BB(\lneg\F)>,
and not $\lneg\BB(F)$,
because otherwise super logic programs would not be rational knowledge bases
(see below).
Then important properties would not hold.}.
Consequently, all of our
results apply to the original AELB as well.  First, let us briefly
recall the basic definitions of AELB.

\subsection{Syntax}

Consider a fixed propositional language $\Lang$ with
standard connectives
($ \lor $, $ \land $, $ \lnot, \ \to $, $\gets,\ \leftrightarrow)$ and the
propositional letters \emph{true} and \emph{false}.  We denote the set
of its propositions by~$\At$.  Extend the language $\Lang$
to a propositional modal language $\LNot$ by augmenting
it with a modal operator $\Not $, called the {\em default negation}
operator.  The formulae of the form $\Not F$, where $F$ is an
arbitrary formula of $\LNot$, are called {\em default
negation atoms} or just {\em default atoms} and are considered to be
atomic formulae in the extended propositional modal language
$\LNot$.  The formulae of the original language
$\Lang$ are called {\em objective},
and the elements of $\At$ are called \emph{objective atoms}.
Any propositional theory in
the modal language $\LNot$ will be called a
\emph{non-monotonic knowledge base} (or ``knowledge base'' for short):

\begin{definition}[Non-monotonic Knowledge Bases]
\label{def-kb}
\mbox{}\\
By a {\em non-monotonic knowledge base} we mean an
arbitrary (possibly infinite) theory in the propositional language $\LNot$.
By using standard logical equivalences,
the theory can be transformed into a set of clauses of the form:
\[ 
\begin{array}{@{}l@{}}
B_1\land \ldots \land B_m \land {\Not}{G_1}\land \ldots \land {\Not}{G_k}\\
\hspace*{2cm} \to A_1\lor \ldots \lor A_l
\lor {\Not}{F_1}\lor \ldots \lor {\Not}{F_n}
\end{array}\] 
where $m,n,k,l \geq 0$, $A_i$'s and $B_i$'s are objective atoms
and $F_i$'s and $G_i$'s are arbitrary formulae of $\LNot$.

By an {\em affirmative} knowledge base we mean any such theory all of
whose clauses satisfy the condition that $l \not= 0$.

By a {\em rational\/} knowledge base we mean any such theory all of whose
clauses satisfy the condition that $n=0$. 
\end{definition} 
 
In other words, \emph{affirmative} knowledge bases are precisely those
theories that satisfy the condition that all of their clauses contain
at least one \emph{objective} atom in their heads. On the other hand,
\emph{rational} knowledge bases are those theories that do not contain
any default atoms ${ {\Not}F_i }$ in heads of their clauses.  Observe
that arbitrarily deep level of \emph{nested default negations} is
allowed in the language~$\LNot$.

\emph{Super logic programs}
(also briefly called ``super programs'')
are a subclass of non-monotonic knowledge bases.
There are three restrictions:
\begin{enumerate}
\item
Only rational knowledge bases are allowed
(i.e.~no default negation in the head).
\item
Nested default negation is excluded.
\item
Default negation can only be applied to positive formulas,
e.g.~${\Not}(\lneg p)$ and ${\Not}(p\rightarrow q)$ cannot be used
in super logic programs.
\end{enumerate}


\begin{definition}[Super Logic Programs]  \mbox{}\\
  A Super Logic Program is a non-monotonic knowledge base consisting
  of (possibly infinitely many) super-clauses of the form:
\[
      F \ {\gets}\  G \land \Not  H
\]
where ${ F }$, ${ G }$ and ${ H }$ are arbitrary positive objective formulae
(i.e.~formulae that can be represented as a disjunction of conjunctions of
      atoms).
\end{definition}

In Proposition~\ref{prop:slp},
it will be shown that simpler clauses can be equivalently used.
However,
one does not need a rule/clause form,
but any formula can be permitted that is equivalent to such clauses.
Our current implementation accepts all formulae
that satisfy the following two conditions:
\begin{enumerate}
\item
inside ${\Not}$ only $\lor$, $\land$, and objective atoms are used,
and
\item
${\Not}$ does not appear in positive context:
\begin{itemize}
\item
An atom~$A$ (objective atom or default negation atom)
appears in the propositional formula~$A$ in positive context.
\item
If $A$ appears in~$F$ in positive context,
it also appears in $F\land G$, $G\land F$, $F\lor G$, $G\lor F$,
$F\leftarrow G$, $G\rightarrow F$, $F\leftrightarrow G$, $G\leftrightarrow F$
in positive context
(where $G$ is any formula).
The same holds for ``positive'' replaced by ``negative''.
\item
If $A$ appears in $F$ in positive context,
it appears in $\lneg F$, $F\rightarrow G$, $G\leftarrow F$,
$F\leftrightarrow G$, $G\leftrightarrow F$ in negative context
(where $G$ is any formula).
The same holds with ``positive'' and ``negative'' exchanged.
\end{itemize}
\end{enumerate}

Clearly the class of super-programs contains all (monotonic)
propositional theories and is syntactically significantly broader than
the class of normal logic programs. In fact, it is somewhat broader
than the class of programs usually referred to as \emph{disjunctive
  logic programs} because:
\begin{enumerate}

\item It allows constraints, i.e., headless rules. In particular it
  allows the addition of \emph{ strong} negation to such programs, as
  shown in Section \ref{sec:strong}, by assuming the \emph{strong
    negation axioms} ${ \ \gets A \land \sneg A \ }$, which themselves are
  program rules (rather than meta-level constraints);

\item It allows premises of the form ${ \Not C }$, where ${ C }$
  is not just an atom but a conjunction of atoms. This proves useful
  when reasoning with default assumptions which themselves are
  \emph{rules}, such as ${ \Not(man \land \sneg human ) }$, which can be
  interpreted as stating that we can assume by default that every man
  is a human (note that ${\sneg human }$ represents strong negation
  of ${ human }$).
\end{enumerate}

\subsection{Implication in the Modal Logic}

The semantics of super logic programs can be seen as an instance of the
semantics of arbitrary nonmonotonic knowledge bases which we introduce now.
AELB assumes the following two simple axiom schemata and one inference
rule describing the arguably obvious properties of default atoms%
\footnote{When  replacing  <\Not(\F)> in these axioms by <\BB(\lneg\F)>,
one gets the axioms of AELB as stated in~\cite{BDP:ainote}.  That paper also
proves the equivalence to the original axioms
of~\cite{P:ael-min,P:ael-minf}.}:

\begin{description}

\item[(CA) Consistency Axiom:]
\mbox{}
\begin{equation}
\label{ax-cons}
\Not(\emph{false}) \text{\ \ and  \ }  \lnot\Not(\emph{true})
\end{equation}

\item[(DA) Distributive Axiom:] \mbox{}\\
\ \ For any formulae $F$ and $G$:
\begin{equation}
\label{ax-norm}
\begin{array}{lcl}
\Not{(F \lor G)} \leftrightarrow \Not F \land  \Not G
\end{array}
\end{equation}

\item[(IR) Invariance Inference Rule:] \mbox{}\\
\ \ For any formulae $F$ and  ${ G }$:
\begin{equation}
\label{rule-necc}
\frac{F \leftrightarrow G}{ \Not F \leftrightarrow \Not G}
\end{equation}

\end{description}

The consistency axiom (CA) states that ${ false }$ is assumed to be
false by default but ${ true }$ is not. The second axiom (DA) states
that default negation \Not\ is distributive with respect to
disjunctions. The invariance inference rule (IR) states that if two
formulae are known to be equivalent then so are their default
negations. In other words, the meaning of ${ \Not F }$ does not depend
on the specific form of the formula ${ F }$, e.g., the formula ${ \Not
  (F \land \lnot F) }$ is equivalent to \mh{ \Not(\emph{false}) } and thus is
true by (CA).

We do not assume the distributive axiom for conjunction:
<\Not{(F \land G)} \leftrightarrow \Not F \lor \Not G>.
This would conflict with the intended meaning of~${\Not}$
(falsity in all minimal models),
see Remark~\ref{rem:drink-drive}.


Of course,
besides the above axioms and inference rule,
also propositional consequences can be used.
The simplest way to define this is via propositional models.
A (propositional) interpretation of~<\LNot>
is a mapping
\[\fullI\colon\;\At\union\setCond{\Not(\F)}{\F\in\LNot}\to
\set{\true,\false},\] i.e.~we simply treat the formulas <\Not(\F)> as
new propositions.  Therefore, the notion of a model carries over from
propositional logic.  A formula <\F\in\LNot> is a propositional
consequence of~<\T\subseteq\LNot> iff for every interpretation~<\fullI>:
<\fullI\models\T\metaThen\fullI\models\F>.  In the examples, we will represent
models by sets of literals showing the truth values of only those
objective and default atoms which are relevant to our considerations.

\begin{definition}[Derivable Formulae]
  \mbox{}\\ 
For any knowledge base $T$ we denote by $
  \CnNot(T)$ the smallest set of formulae of the language $\LNot$
    which contains $T$, all (substitution instances of)
  the axioms (CA) and (DA) and is closed under both propositional
  consequence and the invariance rule (IR).
  
  We say that a formula $F$ is \emph{derivable} from the knowledge
  base ${ T }$ if ${ F }$ belongs to $\CnNot(T)$. We denote this
  fact by \mh{T \vdash_{\Not}F }. A knowledge base $T$ is
  \emph{consistent} if $\CnNot(T)$ is consistent, i.e., if $
  \CnNot(T) \not\vdash_{\Not}$ false.
\end{definition}

The following technical lemma follows by transliteration
from a lemma stated in~\cite{BDP:ainote} for AELB:


\begin{proposition} \label{prop:tech}
For any knowledge base ${ T }$ and any formula ${ F }$
of $\LNot$:
\begin{enumerate}
\item
$T \vdash_{\Not}  (\Not {F}  \to  \lnot\Not \lnot F)$
\item
If $T \vdash_{\Not} {F}$  then
$T \vdash_{\Not}\Not\lnot {F}$. 
\end{enumerate}
 \end{proposition}
 

Since the operator ${ \Not }$ obeys the \emph{distributive law for
  disjunction} (DA), the default atom ${ \Not H }$
in the super logic program rules
can be replaced by
the conjunction ${ \Not C_1 \land \ldots \land \Not C_n }$ of default atoms ${
  \Not C_i }$, where each of the ${ C_i }$'s is a conjunction of
objective atoms.
Together with standard logical equivalences,
this allows us to establish the following useful fact:

\begin{proposition}  \label{prop:slp}
  A super logic program ${ P }$ can be equivalently defined as a set
  of (possibly infinitely many) clauses of the form:
\[
A_1 \lor \ldots \lor A_k \ {\gets} \ B_1 \land \ldots
\land B_m \land \Not C_1 \land \ldots \land \Not C_n ,
\]
where ${ A_i }$'s and ${ B_i }$'s are objective atoms and ${ C_i
  }$'s are conjunctions of objective atoms.
If ${ k \not= 0 }$, for all clauses of ${ P }$, then the program
is called \emph{affirmative}.
\end{proposition}

\subsection{Intended Meaning of Default Negation: Minimal Models}
\label{sec:intended}

As the name indicates, non-monotonic knowledge bases must be equipped
with a \emph{non-monotonic semantics}. Intuitively this means that we
need to provide a meaning to the default negation atoms $\Not F$. We
want the intended meaning of default atoms $\Not F$ to be based on the
principle of \emph{predicate minimization} (see
\cite{Min82,GelPrzPrz89} and \cite{McC80}):
\[ \begin{array}{c}
\Not F  \text{\ \ holds if\ \ } \lnot F  \text{ is minimally entailed }
\\
\text{\ \
or, equivalently:\ \ }  \\ \Not F  \text{\ \ holds if\ \ } F  \text{ is false in
all minimal models. }
\end{array}
\]
In order to make this intended meaning precise we first have to define
what we mean by a minimal model of a knowledge base.


\begin{definition}[Minimal Models]
  \mbox{}\\
A model $M$ is \emph{smaller} than a model $N $ if
  it contains the same default atoms but has fewer
  objective atoms, i.e.{}
  \[\begin{array}{@{}rcl@{}}
  \setCond{\p\in\At}{M\models p}&\subset&\setCond{\p\in\At}{N\models p},\\
  \setCond{F\in\LNot}{M\models\Not(F)}&=&
        \setCond{F\in\LNot}{N\models\Not(F)}.
  \end{array}\]
  By a \emph{minimal model} of a knowledge base ${T}$ we mean a model
  $M$ of ${T}$ with the property that there is \emph{no} smaller model
  $N$ of ${T}$. If a formula $F$ is true in all minimal models of
  ${T}$ then we write $ {T}\ \models _{\min }\ F $ and say that $F$ is
  \emph{minimally entailed} by ${T}$.
\end{definition}

In other words, minimal models are obtained by first assigning
\emph{arbitrary }truth values to the default atoms and then
\emph{minimizing} the objective atoms (see also~\cite{YouYua93auto}). For
readers familiar with \emph{circumscription}, this means that we are
considering circumscription $CIRC({T};\At)$ of the knowledge
base ${T}$ in which objective atoms are minimized while the default
atoms $\Not F\,$ are fixed, i.e., $ T\ \models _{\min }F \ \equiv \
CIRC(T;\At)\models F.$

\begin{example}
\label{ex:broken}Consider the following simple knowledge base $T$:
\[
\begin{array}{lll}
 & \to &\Car\ruleEnd\\
\Car \land {\Not}{\Broken} & \to & \Runs\ruleEnd
\end{array}
\]
Let us prove that ${ T }$ minimally entails ${ \lnot \Broken }$, i.e.,
$T\models_{min} \lnot \Broken $. Indeed, in order to find minimal models of $T$
we need to assign an \emph{arbitrary} truth value to the only default
atom ${\Not}\Broken$, and then \emph{minimize} the objective atoms $
\Broken, \ \Car$ and $\Runs $.  We easily see that $T$ has the following
two minimal models (truth values of the remaining belief atoms are
irrelevant and are therefore omitted): $$
\begin{array}{lcl}
M_1 & = & \{ {\Not}{\Broken},\ \Car, \ Runs, \ \lnot \Broken 
\}; \\
M_2 & = & \{ \lnot{\Not}{\Broken},\ \Car, \ \lnot \Runs, \ \lnot 
\Broken \}.
\end{array}
$$ Since in both of them $ \Car $ is true, and $ \Broken $ is
false, we deduce that $T\models_{min} \Car $ and $T\models_{min} \lnot
\Broken $. 
\end{example}

\subsection{Static Expansions}

The semantics of arbitrary knowledge bases is defined
by means of static expansions.
The characterizations given in Sections~\ref{sec:fixed} and~\ref{sec:model}
show that for the subclass of super logic programs,
simpler definitions are possible.
However,
in order to prove the equivalence
(and to appreciate the simplication),
we first need to repeat the original definition
(adepted to use ${\Not}$ instead of $\BB$).

As in Moore's Autoepistemic Logic, the intended meaning of default
negation atoms in non-monotonic knowledge bases is enforced by
defining suitable expansions of such databases.

\begin{definition}[Static Expansions of Knowledge Bases]\mbox{}\\
A knowledge base $ T^\diamond $ is called a {\em static
  expansion} of a knowledge base $T $ if it satisfies the following
fixed-point equation:
$$ T^\diamond = \CnNot\parenB{T\ \cup \{{{\Not}F\/}: T^\diamond
  \models_{\min } {\lnot F}\}}, $$ where $F$ ranges over all formulae
  of $\LNot$. 
\end{definition}

A static expansion $ T^\diamond $ of $T$ must therefore coincide with the
database obtained by: {(i)} adding to \mh{T} the default negation
${\Not}F$ of every formula \mh{F} that is false in all minimal models
of $ T^\diamond $, and, {(ii)} closing the resulting database under the
consequence operator \mh{Cn_{\Not}}.


 

As the following proposition shows,
the definition of static expansions enforces
the intended meaning of default atoms.
The implication ``$\Leftarrow$'' is a direct consequence of the definition,
the direction ``$\Rightarrow$'' holds for rational knowledge bases%
\footnote{In non-rational knowledge bases,
the user can explicitly assert default negation atoms~${\Not}F$,
even when $F$ is not minimally entailed.}
and thus super logic programs.
The proposition is proven in~\cite{P:ael-min,P:ael-minf}:

\begin{proposition}[Semantics of Default Negation]%
\label{th:converse}
\mbox{}\\
  Let ${ T^\diamond }$ be
  a static expansion of a \emph{rational} knowledge base ${ T }$. For
  any formula ${ F }$ of $\LNot$ we have:
\[
 T^\diamond \models {\Not}F   \text{\ \ \ \ iff \ \ }      T^\diamond \models
_{\min} \lnot F .  
\label{eqn:converse} 
\]
\end{proposition}

It turns out that every knowledge base $T$ has the \emph{least} (in
the sense of set-theoretic inclusion) static expansion $ \overline{T}
$ which has an \emph{iterative} definition as the \emph{least}
\emph{fixed point} of the monotonic\footnote{Strictly speaking the
  operator is only monotone \cite{P:ael-min,P:ael-minf} on the lattice
  of all theories of the form $\CnNot(T \cup N) $,
where $N$ is a set of default atoms.  See also Lemma~\ref{lemTbel}.} default
closure operator:
\[
\Psi_{T}(S)=
        \CnNot\parenB{T \cup \{{{\Not}F\/}:S \ \models_{\min } {\lnot F}\}},
\]
where ${S}$ is an arbitrary knowledge base and the $ F $'s range over
all formulae of $\LNot$.  The following proposition is a
transliteration of a theorem from
\cite{P:ael-min,P:ael-minf,P:static}.  In Section~\ref{sec:fixed} we
will show that it is sufficient to consider only formulas~<\F> of a
special from in~<\Psi_{T}(S)> when we use it for super logic programs.

\begin{proposition}[Least Static Expansion]
\label{th-lst} \mbox{} \\
Every knowledge base $T$ has the \emph{least} static expansion,
namely, the least fixed point $ \overline{T}$ of the monotonic default
closure operator $ \Psi _T$.

More precisely, the least static expansion $ \overline{T}$ of $T$ can
be constructed as follows. Let $T^0 = T$ and suppose that $ T^\alpha $ has
already been defined for any ordinal number $ \alpha \lt\beta $. If $ \beta $$=
\alpha +1$ is a successor ordinal then define: $$
T^{\alpha+1} = \Psi _T(T^\alpha )
=_{def}
\CnNot\parenB{\ T \cup \{{\Not}F: T^\alpha \ \models _{\min }\ \lnot F\}\ 
  }, $$
where $F$ ranges over all formulae in $\LNot$.
Else, if $ \beta $ is a limit ordinal then define $ T^\beta = \bigcup_{\alpha \lt\beta
  }T^\alpha$.  The sequence $\{T^\alpha \}$ is monotonically increasing and
has a limit $ \overline{ T} = T^\lambda = \Psi _T(T^\lambda)$, for
some ordinal $ \lambda $.
\end{proposition}

We were able to show in~\cite{BDP:ainote} that for finite knowledge
bases, the fixed point is in fact already reached after the first
iteration of the default closure operator $\Psi _T$.  In other words,
$\overline{T} = T^1$.

Observe that the {\em least} static expansion $ \overline{T} $ of $T$
contains those and only those formulae which are true in {\em all} static
expansions of $T$. It constitutes the so called \emph{static completion}
of a knowledge base $T$.

\begin{definition}[Static Completion and Static Semantics]
\mbox{}\\
  The least static expansion $\overline{T}$ of a knowledge database
  $T$ is called the \emph{static completion} of $T$. It describes
  the \emph{static semantics} of a knowledge base ${ T }$. 
\end{definition}

%

It is time now to discuss some examples. For simplicity, unless
explicitly needed, when describing static expansions we ignore nested
defaults and list only those elements of the expansion that are
``relevant'' to our discussion, thus, for example, skipping such
members of the expansion as ${\Not}(F \land \lnot F)$, ${\Not}\lnot{\Not}(F \land \lnot
F)$, etc.

\begin{example}
\label{ex:broken2} Consider the database discussed already
in Example~\ref{ex:broken}:
\[
\begin{array}{lll}
& & \Car\ruleEnd \\
\Car \land {\Not}{\Broken} & \to & \Runs\ruleEnd
\end{array}
\]
We already established that ${ T }$ minimally entails ${ \lnot \Broken }$.
As a result, the static completion ${ \overline{T} }$ of ${ T }$
contains ${ \Not \Broken }$.  Consequently, as expected, the static
completion ${ \overline{T} }$ of ${ T }$ derives ${ \Not \Broken }$ and
\mh{\Runs} and thus coincides with the perfect model semantics \cite{P:disjsem} of this
stratified program \cite{AptBlaWal88}.
\end{example}

\begin{example}
\label{ex-2}Consider a slightly more complex knowledge base $T$:
\[
\begin{array}{lll}
{\Not}{\Broken} & \to & \Runs\ruleEnd\\
{\Not}{\Fixed} &\to & \Broken\ruleEnd
\end{array}
\]
In order to iteratively compute its static completion $\overline{T}$
we let $T^0=T$. One easily checks that $T^0\models_{min} \lnot
\Fixed$.  Since
\[
T^1=\Psi_T(T^0)=\CnNot\parenB{T\;\cup \;\{{\Not}F: T^0 \models_{\min}
\lnot F \}}, 
\]
it follows that $\Not \Fixed \in T^1$ and therefore $\Broken \in T^1$.
Since,
\[
T^2=\Psi _T(T^1)=
        \CnNot\parenB{T\;\cup \;\{{\Not}F: T^1 \models_{\min} \lnot F \}},
\]
it follows that $\Not\lnot \Broken \in T^2$.
Proposition~\ref{prop:tech} implies $\lnot\Not \Broken \in T^2$ and
thus $T^2 \models _{\min }\lnot \Runs$. Accordingly, since:
\[
T^3=\Psi _T(T^2)=
        \CnNot\parenB{T\;\cup \;\{{\Not}F: T^2 \models_{\min}\lnot F \}},
\]
we infer that $\Not \Runs \in T^3$. As expected, the static completion
$ \overline{T} $ of $T$, which contains \mh{ T^3}, asserts that
the car is considered not to be fixed and therefore broken and thus is
not in a running condition. Again, the resulting semantics coincides
with the perfect model semantics of this stratified program. 
\end{example}

\begin{example}
\label{ex:visit}%
Consider a simple super-program  ${ P }$:
\begin{eqnarray*}
\VisitEurope \lor \VisitAustralia & {\gets}& \ruleEnd \\
\Happy \ & {\gets}&\  \VisitEurope \lor \VisitAustralia\ruleEnd \\
\Bankrupt \ &{\gets}&\   \VisitEurope \land \VisitAustralia\ruleEnd \\
\Prudent \ &{\gets}&\  \Not (\VisitEurope \land \VisitAustralia)\ruleEnd\\
\Disappointed \ &{\gets}&\  \Not (\VisitEurope \lor \VisitAustralia)\ruleEnd
\end{eqnarray*}
Obviously, the answer to the query $\VisitEurope \lor
\VisitAustralia $ must be positive,
while the answer to the query
$\VisitEurope \land \VisitAustralia $ should be negative. As a result, we
expect a positive answer to the queries $ \Happy$ and $\Prudent $ and a
negative answer to the queries $\Bankrupt $ and $\Disappointed$.
This means that default negation interprets disjunctions in an exclusive way,
as usual in disjunctive logic programming approaches based on minimal models.

Observe that the query $\Not (\VisitEurope \land \VisitAustralia)$
intuitively means \emph{``can it be assumed by default that we don't
  visit both Europe and Australia?'' } and thus it is different from
the query $\Not \VisitEurope \land \Not \VisitAustralia $ which says
\emph{``can it be assumed by default that we don't visit Europe and that we don't visit Australia?"}.

It turns out that the static semantics produces precisely the intended
meaning discussed above. Indeed, clearly ${ \Happy }$ must belong to
the static completion of ${ P }$.  It is easy to check that <\lnot
\VisitEurope \lor \lnot \VisitAustralia>, i.e.{}
\[\lnot( \VisitEurope \land \VisitAustralia)\]
holds in all minimal models of the program
\mh{P} and therefore 
\[\Not(\VisitEurope \land \VisitAustralia)\]
must be true in the completion. This proves that ${ \Prudent }$ must hold
in the completion. Moreover, since ${ \Bankrupt }$ is false in all
minimal models ${ \Not \Bankrupt }$ must also belong to the
completion. Finally, since ${ \VisitEurope \lor \VisitAustralia }$ is
true, we infer from Proposition \ref{prop:tech} that
${\Not\lnot(\VisitEurope \lor \VisitAustralia) }$ and thus also ${
  \lnot\Not(\VisitEurope \lor \VisitAustralia) }$ holds. But then ${
  \lnot \Disappointed }$ is minimally entailed,
thus ${ \Not \Disappointed }$ belongs to the completion. 
\end{example}

\begin{remark}
 \label{rem:drink-drive} 
In general, we do not assume the distributive axiom for conjunction:
\[
\Not{(F \land G)} \leftrightarrow \Not F \lor  \Not G.
\]
It is not difficult to prove that for a rational database \mh{T}
and for any two formulae \mh{F} and \mh{G}:
\[
 T^\diamond \models (\Not F \lor  \Not G)    \text{\ \ \ iff \ \ }
     (T^\diamond \models \Not F  \text{\ \ or \ \ }  T^\diamond \models \Not G).
\]
But together with the axiom in question this would mean
\[
 T^\diamond \models \Not (F \land  G)    \text{\ \ \ iff \ \ }
     (T^\diamond \models \Not F  \text{\ \ or \ \ }  T^\diamond \models \Not G).
\]
This is too restrictive.
For example, given the fact that we can assume by default that we do
not drink and drive (i.e., \mh{T^\diamond \models\Not{(\Drink \land \Drive)}})
at the same time, we do not necessarily want to conclude that we can either
assume by default that we don't drink (i.e., \mh{T^\diamond
  \models\Not{(\Drink)}}) or assume by default that we don't drive (i.e.,
\mh{T^\diamond \models\Not{(\Drive)}}).

In rational knowledge bases, we only assume <\Not(\F)> if <\F> is
false in all minimal models.  But it might well be that <\F> is true
in some, and <\G> is true in some, but they are never true together.
So the axiom in question conflicts with the intended meaning
of default negation.
On some evenings we drink, on some we drive, but we never do both.  If
this is all the information we have, <\Not{(\Drink \land \Drive)}> should
be implied, but neither <\Not{(\Drink)}> nor <\Not{(\Drive)}> should follow.

Note, however, that the implication in one direction, namely <\Not{(F
  \land G)} \gets \Not F \lor \Not G>, easily follows from our axioms.  In
some specific applications, the inclusion of the distributive axiom
for conjunction may be justified; in such cases it may simply be added
to the above listed axioms.
\end{remark}

\subsection{Basic Properties of Super Logic Programs}






The next theorem summarizes properties of super programs.  It is
an immediate consequence of results established in
\cite{P:ael-min,P:ael-minf}:

\begin{theorem}[Basic Properties of Super Programs]  \mbox{}
\\ \vspace{-.2in}  \label{th:basic}
\begin{enumerate}
\item
  Let $P$ be any super logic program and let $\overline{P}$
  be its static completion. For any formula $F$,
    $\overline{P} \models \Not F \ \equiv\ \overline{P} \models_{min} \lnot F$.
\item
  Let $P$ be any super logic program and let $\overline{P}$
  be its static completion.
  For any two formulae \mh{F} and \mh{G} holds:
\[
 \overline{P} \models (\Not F \lor  \Not G)    \text{\ \ \ iff \ \ }
 (\overline{P} \models \Not F  \text{\ \ or \ \ } \overline{P} \models \Not G).
\]
\item Every affirmative super program has a consistent static
  completion. In particular, this applies to all disjunctive logic
  programs.

\item For normal logic programs, static semantics coincides with the
  well-founded semantics \cite{vgrs:wfs}.  More precisely, there is a one-to-one
  correspondence between consistent static expansions of a normal
  program and its partial stable models \cite{P:wfstable}. Under this correspondence,
  (total) stable models \cite{GL:stable} (also called answer sets) correspond to those consistent static expansions
  which satisfy the axiom ${ \Not A \lor \Not \lnot A }$, for every
  objective atom ${ A }$.

\item For positive disjunctive logic programs, static semantics
  coincides with the minimal model semantics. 
\end{enumerate}
\end{theorem}



\subsection{Adding Strong Negation to Knowledge Bases}
 \label{sec:strong}
 
 As pointed out in \cite{GL:classic,APP:neg-ful}
 in addition to default negation, ${ \Not F }$, non-monotonic
 reasoning requires another type of negation \mh{\sneg F} which is
 similar to classical negation ${ \lnot F }$ but does not satisfy the law
 of the excluded middle (see \cite{APP:neg-ful} for more information).

 Strong negation \mh{\sneg F} can be easily added to super logic programs
 by:
(1) Extending the original objective language ${\mathcal L}$ by adding
  to it new objective propositional symbols {$ \sneg A$},
  called \emph{strong negation atoms}, for all \mh{A \in At_{\mathcal L}}.
(2) Adding to the super logic program the following {\em strong negation
  clause}, for any strong negation atom \mh{\sneg A} that appears in it:
$A\ \land \ {\sneg A} \ \to \ false$,
which says that $A$ and $ \sneg A$ cannot be both true.
Since the addition of strong negation clauses simply results in a new
super logic program, our the results apply as well to
programs with strong negation.

\section{Fixed Point Characterization of Static Completions}
\label{sec:fixed}

Due to Proposition \ref{prop:slp}, we can consider the language of
super logic programs to be restricted
to the subset~<{\mathcal L}^*_{\Not}\subseteq{\mathcal L}_{\Not}>,
which is the propositional logic over the following set of atoms:
\[
\At \cup \set{\Not E: E \text{\ is a
    conjunction of objective atoms from~<\At>} }.
\]
In particular, the language ${ {\mathcal L}^*_{\Not} }$ does \emph{not}
allow any \emph{nesting} of default negations. In fact, default
negation can be only applied to conjunctions of objective atoms.
Note, however, that we allow also the ``empty conjunction'', so
<\Not(\true)> is contained in~<{\mathcal L}^*_{\Not}>.  Note also that
we can treat <E> essentially as a set --- because of the invariance
inference rule <\Not E_1> and <\Not E_2> must have the same truth
value if <E_1> and <E_2> differ only in the sequence or multiplicity
of the atoms in the conjunction.
In this way,
if $\At$ is finite,
only a finite number of default negation atoms must be considered.

Clearly, in order to answer queries about a super program ${ P }$ we
only need to know which formulae of the restricted language ${
  {\mathcal L}^*_{\Not} }$ belong to the static completion of ${ P }$.
In other words, we only need to compute the restriction ${
  \overline{P}|{\mathcal L}^*_{\Not} }$ of the static completion ${
  \overline{P} }$ to the language ${ {\mathcal L}^*_{\Not} }$.  It
would be nice and computationally a lot more feasible to be able to
compute this restriction without having to first compute the full
completion, which involves arbitrarily deeply nested default negations
and thus is inherently \emph{infinite} even for finite programs. The
following result provides a positive solution to this
problem in the form of a much simplified syntactic fixed point
characterization of static completions. In the next section we provide
yet another solution in the form of a model-theoretic characterization
of static completions.

\begin{theorem}[Fixed-Point Characterization] 
\label{TH:COMPUTE}
\mbox{}\\ 
The restriction of the static completion $
\overline{P} $ of a finite super program ${ P }$ to the language ${
  {\mathcal L}^*_{\Not} }$ can be constructed as follows. Let $P^0=P$
and suppose that $ P^n $ has already been defined for some natural
number ${ n }$.  Define $P^{n+1}$ as follows: $$
\begin{array}{@{}l@{}l@{}}
Cn\parenB{\ P\ \cup\ &\{\Not E_1 \land\ldots\land\Not E_m \to \Not E_0 :\\
&\hspace*{1cm}
P^n\ \models_{\min } \lnot E_1 \land\ldots\land\lnot E_m \to \lnot E_0  \}\ },
\end{array}
$$
where $E_i$'s range over all conjunctions of objective atoms and ${
  Cn }$ denotes the \emph{standard propositional consequence
  operator}.  We allow the special case that ${ E_0 }$ is the empty
conjunction (i.e.,~{true}) and identify ${ \Not{true} }$ with~{false}.

The sequence $ \{ P^n \} $ is monotonically increasing and has a
limit $ P^{n_0} = P^{n_0+1}$, for some natural number ${
  n_0 }$. Moreover, ${P^{n_0} = \overline{P}|{\mathcal L}^*_{\Not} }$,
i.e., ${ P^{\n_0} }$ is the restriction of the static completion of ${
  P }$ to the language ${ {\mathcal L}^*_{\Not} }$.
\end{theorem}

\begin{proof}
Contained in the Appendix~\ref{app:fixpoint}.
\end{proof}

The above theorem represents a considerable simplification over the
original characterization of static completions given in Proposition
\ref{th-lst}. First of all, instead of using the modal consequence
operator ${ Cn_{\Not} }$ the definition uses the \emph{standard }
propositional consequence operator ${ Cn }$.  Moreover, instead of
ranging over the set of all (arbitrarily deeply nested) formulae of
the language ${ {\mathcal L}_{\Not} }$ it involves only conjunctions of
objective atoms. These two simplifications greatly enhance the
implementability of static semantics of finite programs.  On the other
hand, due to the restriction to the language ${ {\mathcal L}^*_{\Not}
  }$, we cannot expect the fixed point to be reached in just one step,
as would be the case with the more powerful operator~<\Psi_{T}(S)>
\cite{BDP:ainote}.

\section{Model-Theoretic Characterization of Static Completions}

\label{sec:model}

\newcounter{savedSecModelTheoretic}
\setcounter{savedSecModelTheoretic}{\value{section}}

In this section, we complement Theorem \ref{TH:COMPUTE} by providing a
\emph{model-theoretic} characterization of static completions of
finite super programs.  More precisely, we characterize models of
static completions restricted to the narrower language~<{\mathcal
  L}^*_{\Not}>.  The resulting characterization was directly used in a
prototype implementation of a query answering interpreter for static
semantics described in the next section.

\begin{definition}[Reduced Interpretations]\mbox{}
\begin{enumerate}
\item Let <\Obj> be the set of all propositional valuations of the
  objective atoms <\At>.
\item Let <\Def> be the set of all propositional valuations of the
  default atoms \mh{\setCond{\Not(\p_1\landUntil\p_\n)}{\p_{\i}\in\At}},
  which interpret <\Not(\true)> (in case <n=0>) as false%
\footnote{Throughout this section,
whenever we mention a sequence \mh{p_1,\ldots,p_n},
we allow \mh{n} to be 0 as well.}.
\item We call an interpretation~<\redI> of the language~<{\mathcal
    L}^*_{\Not}>, i.e.,~a valuation of
  \mh{\At\union\setCond{\Not(\p_1\landUntil\p_\n)}{\p_{\i}\in\At}},
  a reduced interpretation and write it as
  as~<\redI=\objI\union\defI> with <\objI\in\Obj> and <\defI\in\Def>.
\item In order to emphasize the difference, we sometimes call an
  interpretation~<\fullI> of the complete language~<\LNot>,
  i.e.,~a valuation for <\At\union\setCond{\Not(F)}{\F\in{\mathcal
      L}_{\Not}}>, a full interpretation.
\item Given a full interpretation~<\fullI>, we call its restriction
  <\redI=\objI\union\defI> to the language~<{\mathcal L}^*_{\Not}> the
  reduct of~<\fullI>.
\end{enumerate}
\end{definition}

Since an interpretation~<\redI=\objI\union\defI> assigns truth values
to all atoms occurring in a super program, the ``is model of''
relation between such interpretations and super programs is well
defined.  We call <\redI=\objI\union\defI> a minimal model of a super
program~<{\P}> if there is no interpretation~<\objI\in\Obj>
with
\[\setCond{\p\in\At}{\objI'\models\p}
        \;\subset\;\setCond{\p\in\At}{\objI\models\p}\]
such that <\redI'=\objI'\union\defI> is also a model of~<{\P}>.
This is completely compatible with the corresponding notion
for full interpretations.

Our goal in this section is to characterize the reducts of the full
models of the static completion~<\overline{\P}> of a finite super
program~<{\P}>.  In fact, once we have the possible interpretations of
the default atoms, finding the objective parts of the reducts is easy.
The idea how to compute such possible interpretations is closely
related to Kripke structures (which is not surprising for a modal
logic, for more details see Appendix~\ref{appKripke}).  The worlds in
such a Kripke structure are marked with interpretations
of the objective atoms~$\At$, and
the formula~<\Not(\p_1\landUntil\p_\n)> is true in a world~<\w> iff
<\lnot(\p_1\landUntil\p_\n)> is true in all worlds~<\w'> which can be
``seen'' from~<\w>.  Due to the consistency axiom, every world~<\w>
must see at least one world~<\w'>.  The static semantics ensures that
every world~<\w> sees only worlds marked with minimal models.
So given a set~<\objSet> of (objective parts of) minimal models, an
interpretation~<\defI> of the default atoms is possible iff there is
some subset~<\objSet'\subseteq\objSet> (namely the interpretations
in the worlds~<\w'>) such that <\defI\models\Not(\p_1\landUntil\p_\n)> iff
<\objI\models\lnot(\p_1\landUntil\p_\n)> for all~<\objI\in\objSet'>.
Conversely, <\objSet> also depends on the
possible interpretations of the default atoms (since a minimal model
contains an objective part and a default part).  So when we restrict
the possible interpretations for the default negation atoms,
we also get less minimal models.  This obviously results in a fixed-point
computation, formalized by the following operators:


\begin{definition}[Possible Interpretation of Default Atoms]\mbox{}\\
\label{def:modtheoretic}%
Let~<{\P}> be a super program.
\begin{enumerate}
\item The operator <\DefToObj{\P}\colon2^{\Def}\to2^{\Obj}> yields
  objective parts of minimal models given possible interpretations of
  the default atoms.  For every~<\defSet\subseteq\Def>, let
\[
\begin{array}{ll}
\DefToObj{\P}(\defSet)\defEq
 \setCondB{\objI\in\Obj}{
                &\mbox{there is <\defI\in\defSet> such that}\\
                &\mbox{<\redI=\objI\union\defI>
                        is minimal model of~<{\P}>}}.
\end{array}
\]
\item
Let <\objSet\subseteq\Obj> and <\defI\in\Def>.
We call <\defI> given by~<\objSet> iff
for every <p_1\until\p_\n\in\At>:
\[\begin{array}{@{}l@{}}
\defI\models\Not(\p_1\landUntil\p_\n)\\
                \hspace*{2cm}\mbox{<\iff> for every~<\objI\!\in\objSet>:
                        <\objI\models\lnot(\p_1\landUntil\p_\n)>}.
\end{array}\]
\item The operator <\ObjToDef{\P}\colon2^{\Obj}\to2^{\Def}> yields
  possible interpretations of the default atoms, given
  objective parts of minimal models.  For every~<\objSet\subseteq\Obj>, let
\[\begin{array}{@{}l@{}l@{}}
        \ObjToDef{\P}(\objSet)\defEq\setCondB{\defI\in\Def}{
        &\mbox{there is a non-empty~<\objSet'\subseteq\objSet>}\\
        &\mbox{\qquad such that <\defI> is given by~<\objSet'>}\\
        &\mbox{and there is <\objI> such that}\\
        &\mbox{\qquad <\redI=\objI\union\defI> is a model of~<{\P}>.}}.
\end{array}\]
\item
The operator~<\DefToDef{\P}\colon2^{\Def}\to2^{\Def}>
is the composition~<\DefToDef{\P}\defEq\ObjToDef{\P}\circ\DefToObj{\P}>
of~<\DefToObj{\P}> and~<\ObjToDef{\P}>.
\end{enumerate}
\end{definition} 

Note that the additional constraint in the definition of~<\ObjToDef{\P}>
(that it must be possible to extend <\defI> to a model of~<{\P}>)
is trivially satisfied for affirmative programs,
i.e.~programs having at least one head literal in every rule.

Now we are ready to state the main result of this section.

\newcounter{savedThModelTheoretic}
\setcounter{savedThModelTheoretic}{\value{theorem}}

\begin{theorem}[Model-Theoretic Characterization]
\label{TH:MODELTHEORETIC}%
\mbox{}\\
Let \mh{P} be a finite super program:
\begin{enumerate}
\item The operator~<\DefToDef{\P}> is monotone (in the lattice of
  subsets of~<\Def> with the order <\supseteq>) and thus its iteration
  beginning from the set~<\Def> (all interpretations of the default
  atoms, the bottom element of this lattice) has a fixed
  point~<\DefFix>.
\item <\DefFix> consists exactly of all <\defI\in\Def> that
  are default parts of a full model <\fullI> of <\overline{\P}>.
\item A reduced interpretation~<\objI\union\defI> is a reduct of a
  full model~<\fullI> of~<\overline{\P}> iff <\defI\in\DefFix> and
  <\objI\union\defI\models{\P}>.
\end{enumerate}
\end{theorem}

\begin{proof}
Contained in the Appendix~\ref{app:modeltheoretic}.
\end{proof}

 \begin{example}
\label{ex:modtheoretic}%
Let us consider the following logic program~<{\P}>:
\[\begin{array}{@{}rcl@{}}
p\lor q&\lif&\Not\,r.\\
q&\lif&\Not\,q.\\
r&\lif& q.
\end{array}\]
Until the last step of the iteration, it suffices to consider only
those default atoms which occur in~<{\P}> and thus have influence on the
objective atoms.  But for the sake of demonstration, we start with the
set~<\Def> containing all eight possible valuations of the three
default atoms~<\Not\,p>, <\Not\,q>, and~<\Not\,r> (however, we do not
consider conjunctive default atoms here).  These default
interpretations can be extended to the minimal reduced models listed
in the following table.  Note that models numbered 6 to~10 are equal
to models numbered 1 to~5, except that <\Not\,p> is true in them (this
shows that <\Not\,p> is really superfluous at this stage).
\[\begin{array}{@{}|r|c|c|c|c|c|c|@{}}
\hline
\mbox{No.}
&\Not\,p
&\Not\,q
&\Not\,r
&p
&q
&r\\
\hline
\hline
 1.&\0 &\0 &\0 &\0 &\0 &\0 \\
 2.&\0 &\0 &\1 &\1 &\0 &\0 \\
 3.&\0 &\0 &\1 &\0 &\1 &\1 \\
 4.&\0 &\1 &\0 &\0 &\1 &\1 \\
 5.&\0 &\1 &\1 &\0 &\1 &\1 \\
 6.&\1 &\0 &\0 &\0 &\0 &\0 \\
 7.&\1 &\0 &\1 &\1 &\0 &\0 \\
 8.&\1 &\0 &\1 &\0 &\1 &\1 \\
 9.&\1 &\1 &\0 &\0 &\1 &\1 \\
10.&\1 &\1 &\1 &\0 &\1 &\1 \\
\hline
\end{array}\]
For instance, when <\Not\,q> and <\Not\,r> are both interpreted as
false, <p>, <q> and <r> are false in a minimal model.  If <\Not\,q> is
interpreted as true, <q> and <r> are true and <p> is false by
minimality.  If <\Not\,q> is false and <\Not\,r> is true, there are
two possibilities for a minimal model: either <p> can be true, or <q>
and <r> together.  So there are only three possible valuations for the
objective parts of minimal models, i.e.~<\DefToObj{\P}(\Def)> is:
\[\begin{array}{@{}|c|c|c|@{}}
\hline
p
&q
&r\\
\hline
\hline
 \0 &\0 &\0\\
 \1 &\0 &\0\\
 \0 &\1 &\1\\
\hline
\end{array}\]
Note that although there is a minimal model in which <p>, <q>,
and~<r> are all false, there can be other minimal models based on
different interpretations of the default atoms.
Now the above three interpretations are the input for the next
application of~<\ObjToDef{\P}>.  Of course, from every minimal objective
interpretation we immediately get a possible default interpretation if
we translate the truth of~<p> to the falsity of~<\Not\,p> and so on.
But we can also combine minimal objective parts conjunctively and let
<\Not\,p> be true only if <\lnot p> is true in all elements of some set of
minimal models.
Thus,
<\ObjToDef{\P}\parenB{\DefToObj{\P}(\Def)}> is:
\[\begin{array}{@{}|c|c|c|@{}}
\hline
\Not\,p
&\Not\,q
&\Not\,r\\
\hline
\hline
  \1 &\1 &\1 \\
  \0 &\1 &\1 \\
  \1 &\0 &\0 \\
  \0 &\0 &\0 \\
\hline
\end{array}\]
This means that only the models numbered~<1,5,6,10> remain possible
given the current knowledge about the defaults.  Their objective parts
are:
\[\begin{array}{@{}|c|c|c|@{}}
\hline
p&q&r\\
\hline
\hline
 \0 &\0 &\0\\
 \0 &\1 &\1\\
\hline
\end{array}\]
Finally, <p> is false in all of these models, so <\Not\,p> must be
assumed, and only the following two valuations of the default atoms
are possible in the fixed point~<\DefFix>:
\[\begin{array}{@{}|c|c|c|@{}}
\hline
\Not\,p
&\Not\,q
&\Not\,r\\
\hline
\hline
  \1 &\1 &\1 \\
  \1 &\0 &\0 \\
\hline
\end{array}\]
The reduced models of the least static expansion <\overline{\P}>
consist therefore of all reduced interpretations that include one of
these two default parts and are models of~<{\P}> itself. 
\end{example}

Obviously, computing the reduct of a full interpretation is easy.
Conversely, in the Appendix, Definition~\ref{defStandardKripke}, we
define a Kripke structure which allows us to extend the reduced
interpretations to full interpretations, which are models of the
static completion.
%
However, let us observe that not all full models of the least static
expansion can be reconstructed in this way. Instead, we obtain only
one representative from every equivalence class with the same reduct.
For example, one can easily verify that the
program~<{\P}\defEq\set{\p\lif\Not \p}> has infinitely many
different full models (see Example~\ref{exInfinite} in the Appendix).

\section{Implementation of the Model-Theoretic Characterization}

\label{sec:computation}

\newcounter{savedSecComputation}
\setcounter{savedSecComputation}{\value{section}}

We implemented a \emph{query-answering interpreter for the static
  semantics} in the class of super programs
based on the above model-theoretic characterization%
\footnote{See {\tt http://www.informatik.uni-giessen.de/staff/brass/slp/}.
Actually,
there are two implementations:
A prototype written in Prolog (1325 lines, 608 lines of code),
and a version with web user interface and better performance
written in C++ (21809 lines, 10201 lines of code).
The source code of both versions is freely available.
The C++ version is still being further developed.}.
The interpreter has a web interface,
so it is not necessary to install it locally
in order to try it.

\subsection{Parsing and Clause Transformation}

As mentioned before,
the interpreter does not require clause form.
One can use the following logical operators:
\begin{center}
\begin{tabular}{@{}|l|l|l|l|l|@{}}
\hline
Name&Notation&Alt.&Priority&Associativity\\
\hline
Classical Negation ($\lneg$)&{\tt{\char126}}&&1&right assiciative\\
Default Negation ($\Not$)&{\tt not}&&1&cannot be nested\\
Conjunction (and, $\land$)&{\tt{\char38}}&{\tt ,} &2&right associative\\
Disjunction (or, $\lor$)&{\tt |}&{\tt ;}  {\tt v}&3&
							right associative\\
Implication (then, $\rightarrow$)&{\tt -{\char62}}&&4&
							not associative\\
Implication (if, $\leftarrow$)&{\tt {\char60}-}&{\tt :-}&4&
							not associative\\
Equivalence (if and only if, $\leftrightarrow$)&{\tt {\char60}-{\char62}}&&4&
							not associative\\
\hline
\end{tabular}
\end{center}
Default negation can only be used in negative context,
and inside default negation,
only disjunction and conjunction are permitted.
As in Prolog,
atoms are sequences of letters, digits, and underscores
that start with a lowercase letter.
Alternatively,
one can use any sequence of characters enclosed in single quotes.
Every formula must end in a full stop ``{\tt .}''.
For instance,
the last line of Example~\ref{ex:visit} would be entered as:
\begin{center}
\begin{tt}
disappointed {\char60}- not(visit{\char95}europe | visit{\char95}australia).
\end{tt}
\end{center}
The first task of the program is to read the input,
check it for syntactical correctness,
and translate it into clauses as shown in Proposition~\ref{prop:slp}.
This is done with standard techniques.
For instance,
the above input formula is internally represented as
\begin{center}
\begin{tt}
disappointed {\char60}- not(visit{\char95}europe),
				not(visit{\char95}australia).
\end{tt}
\end{center}

\subsection{Intelligent Grounding}

As mentioned above,
the current implementation permits rules with variables,
but every variable must appear in a positive body literal
(``allowedness'', ``range-restriction'').
For instance,
the rule
\begin{center}
\begin{tt}
p(X,a) | p(X,b) :- q(1,X), not r(X).
\end{tt}
\end{center}
is syntactically valid,
since {\tt X} appears in {\tt q(1,X)}.
This check is done after the clause transformation,
so ``positive body literal'' is not literally required in the input.
As in Prolog,
variables start with an uppercase letter,
and the anonymous variable ``{\tt {\char95}}'' is also understood.
Predicate arguments can be atoms, integers, or variables.
Function symbols (term/list constructors) are not permitted
in order to ensure that the ground instantiation is finite.

However,
our implementation does not compute the complete ground instantiation
of the input program,
since this will often be prohibitively large.
Instead,
the set of derivable conditional facts is computed:

\begin{definition}[Conditional Fact]\mbox{}\\
Conditional facts are formulae of the form
\[A_1\lor\ldots\lor\A_k\leftarrow {\Not}C_1\land\ldots\land{\Not}C_n,\]
where the $A_i$ are objective atoms (positive ground literals)
the $C_i$ are conjunctions of objective atoms (positive ground literals),
$k\ge0$, and $n\ge0$.
In the following,
we treat head and body of conditional facts as sets of literals,
and write them as ${\cal A}\lif{\cal C}$
with ${\cal A}=\set{A_1\until A_k}$ and
${\cal C}=\set{{\Not}C_1\until{\Not}C_n}$.
\end{definition}

So conditional facts are rules without positive body literals,
and because of the allowedness condition,
they also cannot contain variables.
Conditional facts were introduced in \cite{Bry89,Bry90,DK89a}.

The derivation is done with the hyperresolution operator,
which is a generalization of the standard $T_P$-operator.
In the non-disjunctive case,
the $T_P$-operator applies a rule $A\lif B_1\land\ldots\land B_m$
by finding facts~$B'_1\until B'_m$ that match the body literals,
i.e.~$B_i\sigma=B'_i$ for a ground substitution~$\sigma$,
and then deriving the corresponding instance~$A\sigma$ of the head literal.
This is applied e.g.~for the bottom-up evaluation in deductive databases.
The generalization for conditional facts is as follows:

\begin{definition}[Hyperresolution Operator]\mbox{}\\
Let $P$ be a set of super logic program rules as in Proposition~\ref{prop:slp}
and ${\cal F}$ be a set of conditional facts.
Then $H_P({\cal F})$ is the set of conditional facts
\[\begin{array}{@{}l@{}}
\set{A_1\sigma,\ldots,A_k\sigma}\union
	({\cal A}_1\setminus\set{B_1\sigma})\union\ldots\union
	({\cal A}_m\setminus\set{B_m\sigma})
\;\lif\\
\hspace*{4cm}
\set{{\Not}C_1\sigma,\ldots,{\Not}C_n\sigma}
	\union{\cal C}_1\union\ldots\union{\cal C}_m
\end{array}\]
such that there is a rule
\[A_1\lor\ldots\lor A_k\leftarrow B_1\land\ldots\land B_m\land
		{\Not}C_1\land\ldots\land{\Not}C_n.\]
in~$P$,
conditional facts ${\cal A}_i\lif{\cal C}_i$, $i=1\until m$ in~${\cal F}$,
and a subsitution~$\sigma$ with $B_i\sigma\in{\cal A}_i$.
\end{definition}

In other words,
the positive body literals are matched with any literal in the head%
\footnote{Actually,
one can make the resolvable literal unique,
see~\cite{Bra96}.
However,
this is not yet done in the current implementation.}
of a conditional fact,
and then the remaining head literals are added to the corresponding instance
of the rule head,
as the conditions are added to the corresponding instance
of the negative body literals.
For example,
if the rule
\[p_1(X)\lor p_2(Y)\lif\q_1(a,X)\land q_2(X,Y)\land {\Not}\, r(Y)\]
is matched with
$\underline{q_1(a,a)}\lor s(b)$ and
$t(a)\lor \underline{q_2(a,b)}\lif {\Not}\, u(c)$,
the derived conditional fact is
\[p_1(a)\lor p_2(b)\lor s(b)\lor t(a)\lif {\Not}\,r(b)\land {\Not}\,u(c).\]

Again,
head and body of conditional facts are treated as sets of literals,
so duplicate literals are removed.
Also for the hyperresolution step itself,
the rule is instantiated only as far as there are matching conditional facts.

The hyperresolution operator is applied iteratively
until no new conditional facts can be derived.
Since there are only finately many possible ground literals%
\footnote{As mentioned before, no function symbols are permitted.
Also,
the conjunctions inside default negation are treated
as sets of objective atoms:
The implementation discovers e.g.~that $\Not(p\land q)$
and $\Not(q\land p\land q)$ are really the same.}
and no duplicates are permitted in head or body,
the process is guranteed to terminate.
The importance of the hyperresolution fixpoint is that it
has the same minimal models as the original program:

\newcounter{savedThHyperFix}
\setcounter{savedThHyperFix}{\value{theorem}}

\begin{theorem}\mbox{}\\
\label{TH:HYPERFIX}%
Let $P$ be a super logic program,
${\cal F}_0\defEq\emptyset$,
and ${\cal F}_{i+1}\defEq H_P({\cal F}_i)$,
and $n$ be a natural number such that ${\cal F}_{n+1}={\cal F}_n$.
Then the following holds for all Herbrand interpretations~$I$:
$I$ is a minimal model of the ground instantiation~$P^*$ of~$P$
if and only if $I$ is a minimal model of ${\cal F}_n$.
\end{theorem}

\begin{proof}
See Appendix~\ref{app:computation}.
\end{proof}

Of course,
for the termination it is important that the implementation
discovers if the same conditional fact was derived again.
However,
also non-minimal conditional facts can be eliminated:
A conditional fact ${\cal A}\lif {\cal C}$
is non-minimal if there is another conditional fact ${\cal A}'\lif{\cal C}'$
with ${\cal A}'\subseteq{\cal A}$ and ${\cal C}'\subseteq{\cal C}$,
where at least one inclusion is proper.
Since in this case the non-minimal conditional fact ${\cal A}\lif {\cal C}$
is logically implied by the stronger conditional fact ${\cal A}'\lif{\cal C}'$,
the set of models is not changed if ${\cal A}\lif{\cal C}$ is deleted.

\subsection{Algorithm for Detection for Duplicate and Non-Minimal
	Conditional Facts}

Since this redundancy test is needed often,
it is essential that it is implemented efficiently.
E.g.~\cite{Sei94} noted that the elimination of duplicate and subsumed
conditional facts took more than half of the total running time
of query evaluation.

The subsumption check used in our implementation
when a newly derived conditional fact ${\cal A}\lif{\cal C}$
is inserted into the set~${\cal F}$ runs in the time
\[O(\mbox{number of occurrences of literals from ${\cal A}\lif{\cal C}$
	in ${\cal F}$}).\]
We use the algorithm from~\cite{BD95general}
that is shown in Figure~\ref{fig:nonmin}.
\begin{figure}
\begin{tt}
\begin{tabular}{@{}l@{}}
testno $:=$ 0;\\
${\cal F}$ $:=$ $\emptyset$;\\
~\\
{\bf procedure} insert(${\cal A}\lif{\cal C}$)\\
{\bf begin}\\
\ \ \ \ {\bf if} ${\cal A}=\emptyset$ {\bf and} ${\cal C}=\emptyset$
		{\bf then}\\
\ \ \ \ \ \ \ \ report inconsistency and exit;\\
\ \ \ \ testno = testno + 1;\\
\ \ \ \ {\bf for each} $A\in{\cal A}$ {\bf do}\\
\ \ \ \ \ \ \ \ {\bf for each} ${\cal A}'\lif{\cal C}'\;\in\;{\cal F}$
			{\bf with} $A\in {\cal A}'$ {\bf do}\\
\ \ \ \ \ \ \ \ \ \ \ \ {\bf if} inc{\char95}overlap(${\cal A}'\lif{\cal C}'$,
				$\vert {\cal A}\vert\:+\:\vert{\cal C}\vert$)
				{\bf then}\\
\ \ \ \ \ \ \ \ \ \ \ \ \ \ \ \ {\bf return};
				/* ${\cal A}\lif{\cal C}$ is non-minimal */\\
\ \ \ \ {\bf for each} $C\in{\cal C}$ {\bf do}\\
\ \ \ \ \ \ \ \ {\bf for each} ${\cal A}'\lif{\cal C}'\;\in\;{\cal F}$
			{\bf with} $C\in {\cal C}'$ {\bf do}\\
\ \ \ \ \ \ \ \ \ \ \ \ {\bf if} inc{\char95}overlap(${\cal A}'\lif{\cal C}'$,
				$\vert {\cal A}\vert\:+\:\vert{\cal C}\vert$)
				{\bf then}\\
\ \ \ \ \ \ \ \ \ \ \ \ \ \ \ \ {\bf return};
				/* ${\cal A}\lif{\cal C}$ is non-minimal */\\
\ \ \ \ /* ${\cal A}\lif {\cal C}$ is minimal, insert it: */\\
\ \ \ \ ${\cal F}$ $:=$ ${\cal F} \union\set{{\cal A}\lif{\cal C}}$;\\
\ \ \ \ length[${\cal A}\lif{\cal C}$] $:=$
	$\vert{\cal A}\vert\:+\:\vert{\cal C}\vert$;\\
\ \ \ \ lastset[${\cal A}\lif{\cal C}$] $:=$ $0$;\\
{\bf end};\\
~\\
{\bf function} inc{\char95}overlap(${\cal A}'\lif{\cal C}'$, $l$): {\bf bool}\\
{\bf begin}\\
\ \ \ \ /* Increment or initialize the overlap counter: */\\
\ \ \ \ {\bf if} lastset[${\cal A}'\lif{\cal C}'$] $=$ testno {\bf then}\\
\ \ \ \ \ \ \ \ overlap[${\cal A}'\lif{\cal C}'$] $:=$
			overlap[${\cal A}'\lif{\cal C}'$] $+$ $1$;\\
\ \ \ \ {\bf else}\\
\ \ \ \ \ \ \ \ overlap[${\cal A}'\lif{\cal C}'$] $:=$ $1$;\\
\ \ \ \ \ \ \ \ lastset[${\cal A}'\lif{\cal C}'$] $:=$ testno;\\
\ \ \ \ /* Check for subsumption: */\\
\ \ \ \ {\bf if} overlap[${\cal A}'\lif{\cal C}'$] $=$
		length[${\cal A}'\lif{\cal C}'$] {\bf then}\\
\ \ \ \ \ \ \ \ {\bf return} {\bf true};
		/* ${\cal A}\lif{\cal C}$ is non-minimal */\\
\ \ \ \ {\bf if} overlap[${\cal A}'\lif{\cal C}'$] $=$ $l$ {\bf then}\\
\ \ \ \ \ \ \ \ ${\cal F}$ $:=$
			${\cal F}\setminus\set{{\cal A}'\lif{\cal C}'}$;
		/* ${\cal A}'\lif{\cal C}'$ is non-minimal */\\
\ \ \ \ {\bf return} {\bf false};\\
{\bf end};
\end{tabular}
\end{tt}
\caption{Algorithm for Detecting Duplicate and Non-Minimal Conditional Facts}
\label{fig:nonmin}
\end{figure}

The basic idea is as follows:
We store for each existing conditional fact an overlap counter.
When a new conditional fact ${\cal A}\lif {\cal C}$ is produced,
we access for each literal in ${\cal A}$ and ${\cal C}$
all conditional facts ${\cal A}'\lif{\cal C}'$ that contain the same literal
and increment the overlap counter of these conditional facts.
If the overlap counter (i.e.~the number of common literals)
reaches the length (number of literals) of ${\cal A}'\lif {\cal C}'$,
then ${\cal A}\lif{\cal C}$ is redundant.
Otherwise,
if it is equal to the length of ${\cal A}\lif{\cal C}$,
then ${\cal A}'\lif{\cal C}'$ is redundant.

It would be inefficient if we had to set the overlap counters
of all existing conditional facts back to~$0$
before each redundancy test.
Therefore,
we do a ``lazy initialization'':
Each test is assigned a unique number,
and we store for each conditional fact the value of the counter
at the last time the conditional fact was accessed.

Our implementation also does a seminaive evaluation
by requiring that at least one body literal is matched
with a conditional fact
that was newly derived in the last hyperresolution round.
Although it is still possible that the same conditional fact
is derived more than once,
it is at least never produced twice in the same way.

\subsection{Residual Program}

In the end,
when we apply the model-theoretic characterization
for a program that contains $n$~different default negation literals,
we need to consider $2^n$ possible interpretations of these literals
and compute minimal models for them.
That is obviously very expensive
and is only possible for relatively small~$n$.

Therefore,
our implementation evaluates the simple cases directly
before it starts the expensive algorithm on the remaining difficult cases.
This is done by means of reduction operators
defined and analyzed in~\cite{BraDix96jlp2}).
Positive reduction means that $\Not p$ can be evaluated to true
if $p$ does not occur in any head of a conditional fact:

\begin{definition}[Positive Reduction]\mbox{}\\
A set ${\cal F}_1$ of conditional facts
is transformed by positive reduction into a set~${\cal F}_2$
of conditional facts
if there is an atom~$p$ and a conditional fact
${\cal A}\lif{\cal C}\;\in\;{\cal F}$
such that $p$ does not occur in any head in~${\cal F}$
and
\[{\cal F}_2=\bigl({\cal F}_1\setminus\set{{\cal A}\lif{\cal C}}\bigr)
		\union\set{{\cal A}\lif{\cal C}'},\]
where ${\cal C}'={\cal C}\setminus\set{\Not\,p}$.
\end{definition}

Negative reduction means that $\Not\,p_1\land\ldots\land\Not\,p_k$
can be evaluated to false
if there is an unconditional fact $p_1\lor\ldots\lor p_k\lif\true$:

\begin{definition}[Negative Reduction]\mbox{}\\
A set ${\cal F}_1$ of conditional facts
is transformed by negative reduction into a set~${\cal F}_2$
of conditional facts
if there is  $p_1\lor\ldots\lor p_k\lif\true$ in~${\cal F}_1$
and ${\cal F}_2={\cal F}_1\setminus\set{{\cal A}\lif{\cal C}}$
where $\set{\Not\,p_1,\ldots,\Not\,p_k}\subseteq {\cal C}$.
\end{definition}

For instance,
in Example~\ref{ex:visit} there is an unconditional fact
\[\VisitEurope\lor\VisitAustralia.\]
As mentioned above,
the last rule in that example is internally represented as
\[\Disappointed\lif\Not(\VisitEurope)\land
\Not(\VisitAustralia).\]
It can be deleted,
because its condition can never be true.
In the example,
this also eliminates two default negation atoms,
which means that fewer interpretations
must be considered in the next step of the algorithm.

The current version of our implementation does not evaluate
default negation literals that contain conjunctions of atoms.
Since the reduction step is only an optimization,
this is possible:
negations of conjunctions remain for the general algorithm.
But we of course aim at strengthening the optimizations
in future versions.

Note that applying negative reduction can make positive reduction
applicable,
and vice versa.
The two transformations are applied as long as possible,
the result is called the residual program.
It is uniquely determined,
the exact sequence of applications of the two transformations is not
important.
The reduction in the size of the program
and the number of distinct default negations can be quite significant.
For instance, no default negations remain if the input
program is stratified and non-disjunctive.

Negative reduction can be implemented with a method that is
very similar to the overlap counting
shown above for the elimination of nonminimal clauses.
Positive reduction is implemented by managing for each
objective atom a counter for the number of occurrences
in conditional fact heads,
as well as a list of atoms for which this counter is~0.

Note that positive and negative reduction do change the set of minimal
models,
since even if there is the fact~$p$,
the interpretation of $\Not p$ in a minimal model is arbitrary.
Only the static semantics ensures that $\Not p$ is false in this case.
The following simple cumulation theorem ensures
that positive and negative reduction
as well as other such optimizations
do not change the static semantics:

\newcounter{savedThOptim}
\setcounter{savedThOptim}{\value{theorem}}

\begin{theorem}
\label{th:optim}%
\begin{enumerate}
\item
Let $T$ be a knowledge base
with $T\models_{\min}\lneg F$.
Then $T$ and $T\union\set{\Not F}$ have the same static expansions.
\item
Let $T_1$ and $T_2$ be knowledge bases with
$\CnNot(T_1)=\CnNot(T_2)$.
Then $T_1$ and $T_2$ have the same static expansions.
\end{enumerate}
\end{theorem}

For instance,
consider the program $\set{p\lif\Not q}$.
Since $q$ is false in all minimal models,
we can add $\Not q$ by~(1) without changing the static semantics.
Both formulas together are logically equivalent to~$\set{p, \Not q}$.
By~(2),
this does not change the static semantics.
Finally,
since also $\set{p}$ alone implies minimally~$\lneg q$,
by (1) we get that $\set{p}$ has the same static semantics
as~$\set{p,\Not q}$.
This proves that the given program can indeed be reduced
to~$\set{p}$ (an example of positive reduction).

For an example of negative reduction,
consider the program~$T_1$:
\[\begin{array}{@{}l@{}}
p\lor q.\\
s\lif\Not p\land\Not q\land\Not r.
\end{array}\]
Negative reduction transforms this program into~$T_2=\set{p\lor q}$.
In order to show that this does not change the static semantics,
we apply part~(2) of the theorem.
We must show that $T_2$ is $\CnNot$-equivalent to~$T_1$:
Proposition~\ref{prop:tech}
tells us that $\Not(\lneg( p\lor q))$ is contained in~$\CnNot(T_2)$,
and then we also get that $\lneg\Not(p\lor q)\;\in\;\CnNot(T_2)$
(by Proposition~\ref{prop:tech}, the invariance inference rule~(IR)
and propositional consequences).
By applying the distributive axiom (DA)
$\lneg(\Not p\land \Not q)$ follows.
But now the rule $s\lif\Not p\land\Not q\land\Not r$
is a propositional consequence,
and therefore also contained in~$\CnNot(T_2)$.

\begin{corollary}
Positive and negative reduction do not change
the static semantics of a program.
\end{corollary}

\subsection{Application of the Model-Theoretic Characterization}

We apply the definition of the operators~<\DefToObj{\P}>
and <\ObjToDef{\P}> from Section~\ref{sec:model} quite literally.
Of course,
we use not the input program~$\P$,
but the residual program~${\cal F}$.
\begin{figure}
\begin{tt}
\begin{tabular}{@{}l@{}}
Let CritNeg be the set of all default negation atoms
		that appear in ${\cal F}$;\\
$\objSet$ $:=$ $\emptyset;$
	/* Objective parts of minimal models,
		filled by procedure modgen */\\
~\\
{\bf function} static(${\cal F}$): Set of static interpretations for CritNeg\\
{\bf begin}\\
\ \ \ \ {\bf for each} interpretation $\defI$ of CritNeg {\bf do}\\
\ \ \ \ \ \ \ \ ${\cal D}$ $:=$
	$\setCond{{\cal A}}{{\cal A}\lif{\cal C}\;\in\;{\cal F},\;
			\defI\models{\cal C}}$;\\
\ \ \ \ \ \ \ \ modgen(${\cal D}$);
		/* Inserts minimal models of ${\cal D}$ into $\objSet$ */\\
\ \ \ \ $\defSet$ $:=$ $\emptyset$;\\
\ \ \ \ {\bf for each} interpretation $\defI$ of CritNeg {\bf do}\\
\ \ \ \ \ \ \ \ {\bf if} possible($\defI$, $\objSet$) {\bf then}\\
\ \ \ \ \ \ \ \ \ \ \ \ {\bf if} there is no
				$\false\lif{\cal C}\;\in\;{\cal F}$
				with $\defI\models{\cal C}$ {\bf then}\\
\ \ \ \ \ \ \ \ \ \ \ \ \ \ \ \ $\defSet$ $:=$ $\defSet\union\set{\defI}$;\\
\ \ \ \ Changed $:=$ {\bf true};\\
\ \ \ \ {\bf while} Changed {\bf do}\\
\ \ \ \ \ \ \ \ Changed $:=$ {\bf false};\\
\ \ \ \ \ \ \ \ $\objSet$ $:=$ $\emptyset$;\\
\ \ \ \ \ \ \ \ {\bf for each} $\defI\in\defSet$ {\bf do}\\
\ \ \ \ \ \ \ \ \ \ \ \ ${\cal D}$ $:=$
		$\setCond{{\cal A}}{{\cal A}\lif{\cal C}\;\in\;{\cal F},\;
				\defI\models{\cal C}}$;\\
\ \ \ \ \ \ \ \ \ \ \ \ modgen(${\cal D}$);\\
\ \ \ \ \ \ \ \ {\bf for each} $\defI\in\defSet$ {\bf do}\\
\ \ \ \ \ \ \ \ \ \ \ \ {\bf if} {\bf not} possible($\defI$, $\objSet$)
					{\bf then}\\
\ \ \ \ \ \ \ \ \ \ \ \ \ \ \ \ $\defSet$ $:=$ $\defSet\setminus\set{\defI}$;\\
\ \ \ \ \ \ \ \ \ \ \ \ \ \ \ \ Changed $:=$ {\bf true};\\
\ \ \ \ {\bf return} $\defSet$;\\
{\bf end};\\
~\\
{\bf function} possible($\defI$, $\objSet$): {\bf bool}\\
{\bf begin}\\
\ \ \ \ $I_{\intersect}$ $:=$ CritNeg;\\
\ \ \ \ NotEmpty $:=$ {\bf false};\\
\ \ \ \ {\bf for each} $\objI\in\objSet$ {\bf do}\\
\ \ \ \ \ \ \ \ /* Interpretations are identified
			with the set of true default atoms */\\
\ \ \ \ \ \ \ \ $\defI'$ $:=$ $\setCond{\Not(p_1\land\ldots\land p_n)\in
			\mbox{{\tt CritNeg}}}{\objI\models
				\neg(p_1\land\ldots\land p_n)}$;\\
\ \ \ \ \ \ \ \ {\bf if} $\defI\subseteq\defI'$ {\bf then}\\
\ \ \ \ \ \ \ \ \ \ \ \ $I_{\intersect}$ $:=$
				$I_{\intersect}\intersect\defI'$;\\
\ \ \ \ \ \ \ \ \ \ \ \ NotEmpty $:=$ {\bf true};\\
\ \ \ \ {\bf return} $\defI=I_{\intersect}$ {\bf and} NotEmpty;\\
{\bf end};
\end{tabular}
\end{tt}
\caption{Application of the Model-Theoretic Characterization}
\label{fig:modcharac}
\end{figure}

As mentioned before,
it suffices to consider only those default negation atoms
that occur explicitly in the residual program.
We call these the ``critical'' default negation atoms.
Other default atoms have no influence on the minimal models and we can
always extend a valuation of the critical default atoms
consistently to a valuation of all default atoms.
The objective atoms that occur inside the default negations
in the residual program are called the critical objective atoms.

The algorithm shown in Figure~\ref{fig:modcharac}
computes the static interpretations for the critical default negation atoms,
i.e.~the fixed point~<\DefFix> of~<\DefToDef{{\cal F}}>
(reduced to critical default negation atoms).
The fixpoint computation starts with the set of all possible interpretations
for the crtical default negation atoms.
This is the most expensive part of the algorithm,
so real improvements must attack this problem.
In order to keep at least the memory complexity in reasonable limits
the set $\defSet$ is materialized only after the first filtering step
(appication of~$\DefToDef{{\cal F}}$).
Without this optimization,
the first half of the procedure {\tt static(${\cal F}$)}
could simply be replaced by assigning $\defSet$ all possible interpretations.

Now the fixpoint computation starts
(the {\tt{\bf while}}-loop in Figure~\ref{fig:modcharac}).
In each step,
first the set $\objSet=\DefToObj{{\cal F}}(\defSet)$ of minimal models,
reduced to the objective atoms is computed.
The default negation part of each such minimal model
must be an element of~$\defSet$.

Our implementation computes
$\objSet=\DefToObj{{\cal F}}(\defSet)$ as follows:
For each default negation interpretation in~$\defSet$,
we evaluate the conditions of the conditional facts in the residual program
and get a set~${\cal D}$ of positive disjunctions.
For this,
any minimal model generator can be used.
We use the algorithm explained in the next subsection.
A nice feature of it is that it computes not interpretations
for all objective atoms,
but only for the critical ones,
i.e.~those objective atoms that appear inside default negations
in the residual program.
Minimal model generators proposed in the literature are, e.g.,
\cite{BY96,NS96:jicslp,Niemela96:tab}.

Once we have computed~$\objSet$,
interpretations are eliminated from~$\defSet$
that do not satisfy the condition of~$\ObjToDef{{\cal F}}(\objSet)$.
However, it would be very inefficient to consider all
subsets~<\objSet'\subseteq\objSet>,
as required in Definition~\ref{def:modtheoretic}.
However, in order to check whether a given
interpretation~<\defI> of the default atoms is selected
by~<\DefToDef{{\cal F}}>, only the maximal~<\objSet'> is of interest.
It is the set of all $\objI\in\objSet$ that satisfy,
for every default atom, the condition:
\[\begin{array}{c}
        \mbox{if
          <\defI\models\Not(\p_1\landUntil\p_\n)>,}
        \mbox{ then <\objI\models\lnot(\p_1\landUntil\p_\n)>}.
\end{array}\]
Obviously, the inclusion of other~<\objI> into~<\objSet'> would
immediately destroy the required property,
namely that $\defI$ is the ``intersection''
of all models in~$\objSet'$:
\[\defI\models\Not(\p_1\landUntil\p_\n)\iff
        \mbox{for all $\objI\in\objSet$: }
		\objI\models\lnot(\p_1\landUntil\p_\n).\]
Of course,
it must also be tested that the set~$\objSet'$ is non-empty.
In the procedure {\tt possible} shown in Figure~\ref{fig:modcharac},
the set~$\objSet'$ is not explicitly constructed,
but instead the interpretation~$I_{\intersect}$ is constructed
as the intersection of all elements of~$\objSet'$.

\subsection{Computation of Minimal Models}

In order to compute~$\DefToObj{{\cal F}}(\defSet)$,
we must compute the objective parts of minimal models,
of which the default negation part is one of the interpretations in~$\defSet$.
As explained above,
we do this by looping over the interpretations~$\defI\in\defSet$,
evaluating the conditions~${\cal C}$
of the formulas~${\cal A}\lif{\cal C}\;\in\;{\cal F}$ in~$\defI$,
and computing minimal models of the resulting positive disjunctions
of objective atoms%
\footnote{One ``last minute'' idea was to compute directly minimal models
of~${\cal F}$,
which could be done with an algorithm similar to the one shown
in~Figure~\ref{fig:modgen}.
This might relieve us from the need to consider exponentially many
interpretations,
but depending on the given rules,
it might backtrack over even more different cases.
Improvements in this direction are subject of future research.}.
The algorithm that is used to compute minimal models
of these disjunctions is shown in Figure~\ref{fig:modgen}.
\begin{figure}
\begin{tt}
\begin{tabular}{@{}l@{}}
{\bf procedure} modgen(${\cal D}$)\\
{\bf begin}\\
\ \ \ \ $I$ $:=$ $\emptyset$; /* Completely undefined interpretation */\\
\ \ \ \ modgen{\char95}rec(${\cal D}$, $I$);\\
{\bf end};\\
~\\
{\bf procedure} modgen{\char95}rec(${\cal D}$, $I$)\\
{\bf begin}\\
\ \ \ \ {\bf if} all critical objective atoms have a truth value in $I$
		{\bf then}\\
\ \ \ \ \ \ \ \ $\objSet$ $:=$ $\objSet\union\set{I}$;\\
\ \ \ \ {\bf else}\\
\ \ \ \ \ \ \ \ select a critical objective atom $p$ that is still
		undefined in~$I$;\\
\ \ \ \ \ \ \ \ eliminate non-minimal disjunctions from ${\cal D}$;\\
\ \ \ \ \ \ \ \ {\bf if} $p$ does not appear in a disjunction in ${\cal D}$
			{\bf then}\\
\ \ \ \ \ \ \ \ \ \ \ \ /* $p$ is surely false */\\
\ \ \ \ \ \ \ \ \ \ \ \ modgen{\char95}rec(${\cal D}$,
				$I\union\set{\lneg p}$);\\
\ \ \ \ \ \ \ \ {\bf else} {\bf if} $p$ appears as a non-disjunctive fact
			in ${\cal D}$ {\bf then}\\
\ \ \ \ \ \ \ \ \ \ \ \ /* $p$ is surely true */\\
\ \ \ \ \ \ \ \ \ \ \ \ modgen{\char95}rec(${\cal D}$, $I\union\set{p}$);\\
\ \ \ \ \ \ \ \ {\bf else}\\
\ \ \ \ \ \ \ \ \ \ \ \ /* $p$ can be false or true */\\
\ \ \ \ \ \ \ \ \ \ \ \ ${\cal D'}$ $:=$ $\setCondB{{\cal A}\setminus\set{p}}{
						{\cal A}\in{\cal D}}$;
			/* Remove $p$ from all disjunctions */\\
\ \ \ \ \ \ \ \ \ \ \ \ modgen{\char95}rec(${\cal D}'$,
					$I\union\set{\lneg p}$);\\
\ \ \ \ \ \ \ \ \ \ \ \ {\bf for each} disjunction $p_1\lor\ldots\lor p_{i-1}
			\lor p\lor p_{i+1}\lor\ldots\lor p_n\;\in\;{\cal D}$
				{\bf do}\\
\ \ \ \ \ \ \ \ \ \ \ \ \ \ \ \ $I'$ $:=$ $I\union\set{p,\lneg p_1,\ldots,
					\lneg p_{i-1},\lneg p_{i+1},\ldots,
					\lneg p_n}$;\\
\ \ \ \ \ \ \ \ \ \ \ \ \ \ \ \ ${\cal D'}$ $:=$
			$\setCondB{{\cal A}\setminus
			\set{p_1,\ldots,p_{i-1},p_{i+1},\ldots,p_n}}{
				{\cal A}\in{\cal D}}\;\union\;\set{p}$;\\
\ \ \ \ \ \ \ \ \ \ \ \ \ \ \ \ modgen{\char95}rec(${\cal D}'$, $I'$);\\
{\bf end};
\end{tabular}
\end{tt}
\caption{Minimal Model Generator}
\label{fig:modgen}
\end{figure}
We only compute the truth values of critical objective atoms,
i.e.~objective atoms that appear inside default negations in~${\cal F}$%
\footnote{In the last recursive call,
$I'$ can assign ``false'' to additional objective atoms,
which is required as a reason for a critical atom being true.}.
Of course,
it is important that this partial interpretation
can be extended to a total minimal model.

So let $p$ be a critical objective atom
that is not yet assigned a truth value.
The two simple cases are:
(1) $p$ does not appear in any disjunction in~${\cal D}$:
Then $p$ must be false in every minimal model.
(2) $p$ appears as a (non-disjunctive) fact in~${\cal D}$:
Then $p$ must be true in all minimal models.

The difficult case is when $p$ appears in one or more proper disjunctions.
Then it can be true or false,
and we backtrack over the different possibilities
(the backtracking is simulated with local variables
and ``call by value'' parameters in~Figure~\ref{fig:modgen},
in the actual implementation,
several stacks are used).
The model construction is based on the following theorem:

\newcounter{savedThModGen}
\setcounter{savedThModGen}{\value{theorem}}

\begin{theorem}\mbox{}\\
\label{TH:MODGEN}%
Let ${\cal D}$ be a set of disjunctions of objective atoms,
which does not contain the empty disjunction~$\false$,
and which does not contain two disjunctions~${\cal A}$
and ${\cal A}'$,
such that ${\cal A}\subset{\cal A}'$ (i.e.~${\cal A}'$ is non-minimal).
Let $I$ be a partial interpretation
such that atoms interpreted as false do not appear in~${\cal D}$
and atoms interpreted as true appear as facts in~${\cal D}$.
Let $p$ be an atom that appears in
the proper disjunction
\[p_1\lorUntil p_{i-1}\lor p\lor p_{i+1}\lorUntil p_n\]
(and possibly more such disjunctions).
Then
\begin{enumerate}
\item
There is a minimal model of~${\cal D}$
that extends~$I$ and interprets $p$ as false.
\item
There is a minimal model of~${\cal D}$
that extends~$I$ and interprets~$p$ as true
and $p_1\until p_{i-1},p_{i+1}\until p_n$ as false.
\end{enumerate}
\end{theorem}

\begin{proof}
See Appendix~\ref{app:computation}.
\end{proof}

First,
$p$ can be false.
In order to ensure that the theorem is again applicable for the next atom,
we eliminate $p$ from all disjunctions in~${\cal D}$
(which also might eliminate further disjunctions that now become non-minimal).
Alternatively,
$p$ can be true.
However,
$p$ can only be true in a minimal model
if for one of the disjunctions in which $p$ appears,
all the remaining atoms are false.
This ensures that there is a reason why $p$ must be true.
Again,
the assigned truth values are reflected by changing the set of disjunctions:
$p$ is added as a fact and
$p_1\until p_{i-1},p_{i+1}\until p_n$ are removed from all disjunctions.
This might make disjunctions in~${\cal D}$ non-minimal,
they are removed with the same overlap counting technique
as shown in Figure~\ref{fig:nonmin}.
In this way,
assumed truth values are immediately propagated in the disjunctions,
which often makes truth values of further critical atoms unique.

The algorithm can also be understood as applying a generalization
of Clark's completion to disjunctions:
One treats the disjunction <\p_1\lorUntil\p_\n> like
the $n$ rules
\[\p_{\i}\lif\lnot\p_1\landUntil\lnot\p_{{\i}-1}
	\land\lnot\p_{{\i}+1}\landUntil\lnot\p_\n\]
and then applies the standard completion
(which basically translates $\lif$ to~$\liff$).
Whereas in general,
Clark's completion does not correspond to minimal models,
in this particular case,
the completion enforces exactly the minimal models
\cite{BDP:ainote}.
For instance,
suppose that $p$ appears in the disjunctions $p\lor q$ and $p\lor r\lor s$.
Then the completion contains the formula
\[p\liff\lneg q\lor(\lneg r\land\lneg s).\]
The above model generation algorithm distinguishes now three cases:
\begin{enumerate}
\item
$p$ is false.
Then it is removed from the disjunctions,
so $q$ and $r\lor s$ are assumed.
\item
$p$ is true and $q$ is false.
\item
$p$ is true and $r$ and $s$ are both false.
\end{enumerate}

In summary, nice features of the algorithm are:
\begin{enumerate}
\item
It never runs into dead ends:
All assumed truth values are indeed possible.
\item
No additional minimality test for the generated models is needed.
\item
It can generate partial minimal models,
i.e.~truth value assignments for any subset of the atoms
that can be extended to minimal models.
\item
The space complexity is polynomial in the size of the input disjunctions.
\end{enumerate}
On the negative side,
the algorithm may generate
the same model several times.

\subsection{Query Evaluation}

The implementation can print the fixed point~$\DefFix$
of~$\DefToDef{{\cal F}}$,
i.e.~all static interpretations of the default negation literals.
Normally,
only the truth values of the critical negations are shown,
but one can add further default negation literals
if one wants to see their truth values.

Queries are written in the form
\[\mbox{{\tt ?\kern0.4em p(X), q(X,b), not r(X).}}\]
They are handled by adding a rule
with a special predicate {\tt {\char36}answer}
in the head:
\[\mbox{{\tt {\char36}answer(X) {\char60}- p(X), q(X,b), not r(X).}}\]
Then the hyperresolution fixpoint is computed as usual.
When positive reduction is applied,
and when minimal models are generated,
conditional facts that contain {\tt {\char36}answer} in the head are ignored.
Conditional facts in the residual program
with only {\tt {\char36}answer} in the head
correspond to possible answers.
Their conditions are evaluated in each of the models in~$\DefFix$.
If the condition is true in all $\defI\in\DefFix$,
the corresponding answer is printed.

\section{Related Work}
\label{sec:relwork}

\subsection{Extended Logic Programs by Lloyd and Topor}

The first attempt to generalize the syntax of logic programs is due to
Lloyd and Topor~\cite{LT84,LT85,LT86}.  They allow arbitrary formulae
in bodies of clauses, which is especially important in order to map
database queries and view definitions into logic programming rules.
However, the head of a rule must still be a single atom.  Furthermore,
their semantics differentiates between an atom occurring in the head
and its negation occurring in the body, whereas our semantics is
invariant under logically equivalent transformations.

Lloyd and Topor use Clark's completion for defining the meaning of
default negation, whereas the static semantics extends the
well-founded semantics.  We have not yet treated quantifiers in the
formulae of knowledge bases, but all other normalization
transformations of Lloyd and Topor work as well for our semantics.
This means that if Lloyd and Topor had used the WFS instead of Clark's
completion, they could have come up with a special case of the static
semantics (restricted to formulae of the form <A\lif W>).

\subsection{Disjunctive Stable Semantics}

The disjunctive stable semantics (answer sets) is
the most popular semantics proposed earlier for disjunctive programs
\cite{P:wfstable,GL:classic}.
Its main problem is that it is inconsistent
for some very intuitive programs,
such as:
\begin{eqnarray*}
work & \gets & \Not tired\ruleEnd\\
sleep & \gets & \Not work\ruleEnd\\
tired & \gets & \Not sleep\ruleEnd\\
angry & \gets &  \Not paid, work\ruleEnd\\
paid & \gets & \ruleEnd
\end{eqnarray*}
Here we should at least be able to conclude $paid$ and $\Not angry$,
and the static semantics gives us just that.
In general,
the static semantics is always consistent for disjunctive logic programs.

In addition to the fact that stable semantics is contradictory for
some very simple and natural programs, it also suffers from a number
of {\em structural} problems. In particular, the stable semantics is not
\emph{relevant}, i.e., answering a query does not depend only on the
call graph below that query
(\cite{BraDix96jlp1,Dix93aclassificationtheoryII}), and it is not
\emph{compositional} (or modular), a highly desirable property for
software engineering and KR \cite{Bry96,BugLamMel94,TeuEta96}. The
static semantics has both of these nice structural properties.  Thus
methods for query optimization based on the dependency graph can be
applied and may reduce the overall computation.

Needless to say, we do not claim that the static semantics should
replace the disjunctive stable semantics.  Like it is the case with
normal logic programs, there are some application domains for which
the disjunctive stable semantics seems to be better suited to
represent their intended meaning~\cite{CMMT95}.  However, we do claim
that the static semantics is a very  natural and well-behaved
extension of the well-founded semantics to disjunctive programs and
beyond.  Because of the importance of the well-founded semantics for
normal programs, it is of great importance to find the its proper
extension(s) to more general theories.

\subsection{Well-Founded Circumscriptive Semantics}

Our work is also related to the approach presented in~\cite{YouYua93auto} where the authors defined the {\em well-founded circumscriptive\/} semantics
for disjunctive programs.
They introduced the concept of minimal model
entailment with fixed interpretation of the default atoms
and defined the semantics of~$T$ as the minimal models of~$T$
which satisfy the limit of the following sequence:
$W^0=\emptyset$ and
\[W^{n+1}\defEq W^n\union\setCond{\lneg\Not p}{T\union W^n\models p}
	\union\setCond{\Not p}{T\union W^n\models_{\min} \lneg p}\]
(this is a translation in our own notation).
In the definition of the static fixpoint operator~$\Psi_T$
(see Proposition~\ref{th-lst}),
negations of default negations are not directly assumed.
But in contrast to the well-founded circumscriptive semantics
which only assumes the default negation of propositions,
the static semantics assumes the default negation of arbitrary propositions.
With Proposition~\ref{prop:tech} it follows that
if $T\union\setCond{\Not F}{T^n\models_{\min}\lneg F}$
implies $F$,
then also $\lneg\Not F$ is contained in~$T^{n+1}=\Psi_T(T^n)$.
With this,
it is easy to see that $W^n\subseteq T^n$.
So the default negation part of the well-founded circumscriptive semantics
is weaker than that of the static semantics.
The reason is that the static semantics assumes negations of arbitrary
formulae, not only of atoms.
E.g.~in Theorem~\ref{TH:COMPUTE} it is really needed
that also implications between default negation atoms are assumed.
For instance,
in Example~\ref{ex:modtheoretic},
the static semantics first derives $\Not q\liff\Not r$
and then $\Not p$ follows.
However,
in the well-founded circumscriptive semantics the limit of the sequence~$W^n$
is the empty set:
Nothing becomes known about the default negation literals.

Another difference between the two approaches is that the well-founded
circumscriptive semantics permits only minimal models
of the objective atoms.
The static semantics considers minimal models when deciding
which default negations to assume,
but if one does not use default negation,
one gets standard propositional logic.
For instance,
let $P=\set{p}$.
If also $q$ belongs the language,
the well-founded circumscriptive semantics implies $\lneg q$ and $\Not q$,
whereas the static semantics implies $\Not q$, but not $\lneg q$.
Of course,
both imply $p$ and $\lneg\Not p$.



%





 

\subsection{Disjunctive Logic Programming Systems}

In addition to the DisLoP project in Koblenz, there is a similar
project on disjunctive logic programming at the Theoretical University
of Vienna called~{\tt dlv} (see \cite{eite-etal-97h,eite-etal-98a}). While DisLoP
concentrated on a disjunctive extension of the wellfounded semantics (D-WFS), {\tt dlv}
computes stable models (answer sets) both for the disjuctive and the non-disjunctive case.
It is a knowledge representation system, which offers front-ends to
several advanced KR formalisms.  The kernel language, which extends
disjunctive logic programming by true negation and integrity
constraints, allows for representing complex knowledge base problems
in a highly declarative fashion \cite{eite-etal-98a}. The project also
incorporates modular evaluation techniques along with linguistic
extensions to deal with quantitative information \cite{bucc-etal-97a}.

\subsection{Nested Rules}

Recently, Lifschitz et.~al.~introduced nested expressions in the
heads and bodies of rules (\cite{LifTanTur98}).
Also nested negation as failure is supported,
which is excluded in super logic programs.
Their semantics is based
on answer sets (also called stable models), whereas in our
framework, we build upon the wellfounded semantics.

In \cite{GreLeoSca98} nested rules are also allowed in rule heads.
The authors show that, in terms of expressive power,
they can capture the full second level of the polynomial hierarchy
(which is also true for other approaches).

\subsection{Other Logics}

There are also successful approaches to generalize logic programming
languages by using intuitionistic, linear, or higher order logics
(e.g.~\cite{HMiller94,NMiller90}).  These extensions seem somewhat
orthogonal to our treatment of negation in the context of arbitrary
propositional formulas.

\section{Conclusion}
\label{sec:conclusions}

We introduced the class of super-programs as a subclass of the class
of all non-monotonic knowledge bases. We showed that this class of
programs properly extends the classes of disjunctive logic programs,
logic programs with strong (or ``classical'') negation and arbitrary
propositional theories. We demonstrated that the {semantics} of super
programs constitutes an intuitively natural extension of the semantics
of normal logic programs. When restricted to normal logic programs, it
coincides with the well-founded semantics, and, more generally, it
naturally corresponds to the class of all partial stable models of a
normal program.

Subsequently, we established two characterizations of the
static semantics of finite super-programs, one of which is syntactic
and the other model-theoretic, which turned out to lead to procedural
mechanisms allowing its computation. Due to the restricted nature of
super programs, these characterizations are significantly
simpler than those applicable to arbitrary non-monotonic knowledge
bases.

We used one of these characterizations as a basis for the
implementation of a \emph{query-answering interpreter} for
super-programs which is available on the WWW.  We noted that
while no such computational mechanism can be efficient, due to the
inherent NP-completeness of the problem of computing answers to just
positive disjunctive programs, they can become efficient when
restricted to specific subclasses of programs and queries.  Moreover,
further research may produce more efficient \emph{approximation
  methods}.

The class of non-monotonic knowledge bases, and, in particular, the
class of super programs, constitutes a special case of a much more
expressive non-monotonic formalism called the \emph{Autoepistemic
  Logic of Knowledge and Beliefs}, \emph{AELB}, introduced earlier in
\cite{P:ael-min,P:ael-minf}.  \emph{AELB} isomorphically includes the
well-known non-monotonic formalisms of Moore's \emph{Autoepistemic
  Logic} and McCarthy's \emph{Circumscription}. Via this embedding,
the semantics of super programs is clearly linked to other
well-established non-monotonic formalisms.

The proposed semantic framework for super programs is sufficiently
flexible to allow various application-dependent extensions and
modifications. We have already seen in Theorem \ref{th:basic} that by
assuming an additional axiom we can produce the stable semantics
instead of the well-founded semantics. By adding the distributive
axiom for conjunction we can obtain a semantics that extends the
\emph{disjunctive stationary semantics} of logic programs introduced
in \cite{P:disjsem}. Many other modifications and extensions are
possible including variations of the notion of a minimal model
resulting in \emph{inclusive}, instead of \emph{exclusive},
interpretation of disjunctions \cite{P:static}.


\begin{acks}
  The authors would like to express their deep appreciation to Luis
  Moniz Pereira and Jose Alferes for their helpful comments. The
  authors are also greatly indebted to the anonymous reviewers
  for their detailed and very insightful comments,
  which have lead to a significant improvement of this article.
\end{acks}


\bibliographystyle{esub2acm}
\bibliography{super-minimal}
\appendix

\section{Kripke Models of Static Expansions}
\label{appKripke}

Before we prove the model-theoretic characterization
(Theorem~\ref{TH:MODELTHEORETIC}), we prove here a Theorem which easily
allows us to construct models of the least static expansion from
Kripke structures.  It is used as a lemma in the proof to
Theorem~\ref{TH:MODELTHEORETIC}, but it is of its own interest.
Although we later need only super programs and reduced models, we
allow in this section arbitrary belief theories and consider full
models.

In order to be precise and self-contained,
let us briefly repeat the definition of Kripke structures~\cite{MarTru93}.
Since our axioms entail the normality axiom,
it suffices to consider normal Kripke structures:

\begin{definition}[Kripke Structure]
\mbox{}\\
A (normal) Kripke structure is a triple~<\K=\struct{\W,\R,\V}> consisting of
\begin{enumerate}
\item
a non-empty set~<\W>, the elements~<\w\in\W> are called worlds,
\item
a relation~<\R\subseteq\W\times\W>, the ``visibility relation''
(if <\R(\w,\w')> we say that world~<\w> sees world~<\w'>), and
\item
a mapping~<\V\colon\W\to\Obj>,
which assigns to every world~<\w>
a valuation~<\objI=\V(\w)> of the objective atoms~<\At>.
\end{enumerate}
\end{definition}

\begin{definition}[Truth of Formulas in Worlds]
\mbox{}\\
The validity of a formula~<\F> in a world~<\w> given Kripke
structure~<\K=\struct{\W,\R,\V}> is defined by
\begin{enumerate}
\item
If <\F> is an objective atom (proposition)~<\p\in\At>,
then <(\K,\w)\models\F\defIff\V(\w)\models\p>.
\item
If <\F> is a negation~<\lnot\G>,
then <(\K,\w)\models\F\defIff(\K,\w)\not\models\G>.
\item
If <\F> is a disjunction~<\G_1\lor\G_2>,
then
\[(\K,\w)\models\F\defIff(\K,\w)\models\G_1\mbox{ or }(\K,\w)\models\G_2\]
(and further propositional connectives as usual).
\item
If <\F> is a default negation atom~<\Not(\G)>, then
\[(\K,\w)\models\F\defIff\mbox{for all <\w'\in\W> with <\R(\w,\w')>: }
(\K,\w')\not\models\G.\]
\end{enumerate}
\end{definition}

Now given such a Kripke structure~<\K>, we get from every world~<\w> a
propositional interpretation~<\fullI=\K(\w)> of~<\LNot>, i.e.~a
valuation of <\At\union\setCondB{\Not(\F)}{\F\in\LNot}>: We simply make
an objective or belief atom~<\A> true in~<\fullI> iff <(\K,\w)\models\A>.
Since the propositional connectives are defined in a Kripke structure
like in standard propositional logic, we obviously have
<\fullI\models\F\iff(\K,\w)\models\F> for all~<\F\in\LNot>.

\begin{theorem}[Kripke Structures Yield Static Expansions]
\label{thKripke}%
\mbox{}\\
Let <\T> be an arbitrary belief theory and <\K=\struct{\W,\R,\V}> be a
Kripke structure satisfying
\begin{enumerate}
\item For every~<\w\in\W>, there is a~<\w'\in\W> with <\R(\w,\w')>
  (consistency).
\item For every~<\w\in\W>, <(\K,\w)\models\T>, i.e.~the
  interpretation~<\fullI=\K(\w)> is a model of~<\T>.
\item For every~<\w,\w'\in\W> with~<\R(\w,\w')>, the
  interpretation~<\fullI=\K(\w')> is a minimal model of~<\T>
  (``only minimal models are seen'').
\end{enumerate}
Then <T^\diamond\defEq\setCond{\F\in\LNot}{\mbox{for every <\w\in\W>:
    <(\K,\w)\models\F>}}> is a static expansion of~<\T>.
\end{theorem}

\begin{proof}
We have to show that
$T^\diamond = Cn_{\Not}\parenB{T\ \cup \{{{\Not}F\/}: T^\diamond
  \models_{\min } {\lnot F}\}}$. The fact that $T\subset T^\diamond$ follows immediately from
  the second assumption. 
First we prove that $T^\diamond$  is closed under $Cn_{\Not}$:
\begin{enumerate}
\item Consistency Axioms: Let any <\w\in\W> be given.  The first part
  <(K,\w)\models\Not(\false)> is trivial, because for any <\w'\in\W> we have
  <(\K,\w')\not\models\false>.  Second, we have to show <(\K,\w)\models\lnot\Not(\true)>.
  By the first requirement of the theorem, there is~<\w'\in\W>
  with~<\R(\w,\w')>.  Now <(\K,\w')\models\true>, therefore
  <(\K,\w)\not\models\Not(\true)>, i.e.~<(\K,\w)\models\lnot\Not(\true)>.
\item Distributive Axiom: We have to show that for all
  formulas~<\F,\G\in\LNot> and all <\w\in\W> that the distributive axiom
  holds in~<\w>: <(\K,\w)\models \Not(\F\lor\G)\;\liff\;Not(F)\land\Not(\G)>.
  This follows simply from applying the definitions:
  <(\K,\w)\models\Not(\F\lor\G)> <\iff> for all <\w'\in\W> with~<\R(\w,\w')>:
  <(\K,\w')\not\models\F\lor\G> <\iff>for all <\w'\in\W> with~<\R(\w,\w')>:
  <(\K,\w')\not\models\F> and <(\K,\w')\not\models\G> <\iff> <(\K,\w)\models\Not(F)> and
  <(K,\w)\models\Not(G)>, <\iff> <(\K,\w)\models\Not(F)\land\Not(\G)>.
\item Invariance Inference Rule: Let <\F\liff\G\in\T^\diamond>, i.e.~for
  every~<\w\in\W> we have <(\K,\w)\models\F\iff(\K,\w)\models\G>.  But then also
  <(\K,\w)\models\Not(\F\liff\G)> holds for every~<\w\in\W> since
  <(K,\w')\not\models\F\iff(\K,\w')\not\models\G> holds for every~<\w'\in\W>
  with~<\R(\w,\w')>.
\item Closure under propositional consequences: Let <\F\in\LNot> a
  formula which is a propositional consequence
  of~<\F_1\until\F_\n\in\T^\diamond>.  Now <\F_{\i}\in\T^\diamond> means that
  <(\K,\w)\models\F_{\i}> for every~<\w\in\W>.  But the formulas valid in one
  world are closed under propositional consequences, since the meaning
  of the propositional connectives is defined as in the standard case.
  So we get <(\K,\w)\models\F>, and thus <\F\in\T^\diamond>.
\end{enumerate}
Now suppose that <\T^\diamond\models_{\rm min}\lnot\F>,
  i.e.~<\F> is false in all minimal models of~<\T^\diamond>.
  We have to show that <\Not(\F)\in\T^\diamond>,
  i.e.~<(\K,\w)\models\Not(\F)> for every~<\w\in\W>.
  So we have to show <(\K,\w')\not\models\F>
  for every~<\w'\in\W> with <\R(\w,\w')>.
  
  Let such a~<\w'> be given, and let <\fullI=\K(\w')>.  By the last
  condition of the theorem, we know that <\fullI> is a minimal model
  of~<\T>.  By construction, it is also a model of~<\T^\diamond>, and since
  <\T\subseteq\T^\diamond>, there can be no smaller model.
  Thus, <\fullI> is a minimal model of~<\T^\diamond> and therefore
  satisfies~<\lnot\F>, i.e.~<(\K,\w')\not\models\F>.
  
  Thus we have shown that $Cn_{\Not}\parenB{T\ \cup \{{{\Not}F\/}: T^\diamond
  \models_{\min } {\lnot F}\}}\subseteq T^\diamond$. The converse follows from
  assumption (3) and (4) of the preceding definition.
\end{proof}

This theorem gives us a simple way to construct models of the least
static expansion~<\overline{\T}>: Obviously, for every~<\w\in\W>, the
interpretation~<\fullI=\K(\w)> is a model of~<T^\diamond>.  But the least
static expansion is a subset of every other static expansion,
i.e.~<\overline{\T}\subseteq\T^\diamond>, and therefore we
have~<\fullI\models\overline{\T}>.

\begin{corollary}
\label{corKripke}%
If a Kripke structure~<\K=\struct{\W,\R,\V}> satisfies the conditions
of Theorem~\ref{thKripke}, then for every <\w\in\W>, the
interpretation~<\fullI=\K(\w)> is a model of the least static
expansion~<\overline{\T}>.
\end{corollary}

\begin{example}
Let us consider the knowledge base of Example~\ref{ex-2}:
\[
\begin{array}{lll}
{\Not}{\Broken} & \to & \Runs\ruleEnd\\
{\Not}{\Fixed} &\to & \Broken\ruleEnd
\end{array}
\]
Here, we can construct a Kripke model~<\K> with only one world~<\w>
with the valuation~<\objI=\set{\lnot{\Fixed},\:{\Broken},\:\lnot{\Runs}}>.  We
let this world ``see'' itself, i.e.~<\R\defEq\setB{(\w,\w)}>.
Then <{\Not}{\Fixed}> and <{\Not}{\Runs}> are true in <(\K,\w)>, but
<{\Not}{\Broken}> is false.

It is, however, also possible to add a world~<\w'> with a non-minimal
valuation~<\objI'=\set{{\Fixed},\:{\Broken},\:{\Runs}}>.  Both worlds can
only see~<\w>, i.e.~<\R\defEq\setB{(\w,\w),\:(\w',\w)}>, because all seen
worlds must be minimal models.  This example shows that the static
semantics does not imply <\lnot{\Fixed}>, but it of course implies
<{\Not}{\Fixed}>.  So the non-monotonic negation is cleanly separated
from the classical negation.
\end{example}

\begin{example}
\label{exInfinite}%
Let us consider the theory~<{\P}\defEq\set{\p\lif\Not(\p)}> corresponding
to a well-known logic program.  We claimed in Section~\ref{sec:model}
that the least static expansion~<\overline{\P}> of this program has
infinitely many different models.  But let us first look a Kripke
structure which generates only two models:
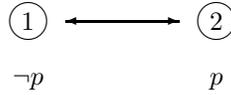
\begin{center}
\setlength{\unitlength}{1mm}
\begin{picture}(35,22)(0,0)
\put(5,13){\makebox(0,0){1}}
\put(5,13){\circle{5}}
\put(5,5){\makebox(0,0){<\lnot\p>}}
\put(30,13){\makebox(0,0){2}}
\put(30,13){\circle{5}}
\put(30,5){\makebox(0,0){<\p>}}
\put(11,13){\vector(1,0){14}}
\put(24,13){\vector(-1,0){14}}
\end{picture}
\end{center}
I.e.~the set of worlds~<\W> is~<\set{1,2}>, the visibility
relation~<\R> is~<\setB{(1,2),\;(2,1)}> (each world can see the other
one, but not itself), and <\p> is false in world~1 and true in
world~2.  In world~<1>, <\Not(\p)> is false, and in world~<2>,
it is true.

But there are also quite different Kripke models.  The construction in
Section~\ref{sec:model} will yield:
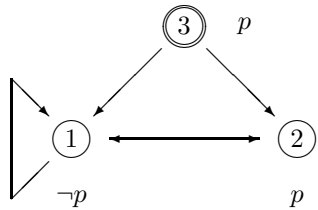
\begin{center}
\setlength{\unitlength}{1mm}
\begin{picture}(48,37)(0,0)
\put(13,13){\makebox(0,0){1}}
\put(13,13){\circle{5}}
\put(13,5){\makebox(0,0){<\lnot\p>}}
\put(19,13){\vector(1,0){19}}
\put(10,10){\line(-1,-1){5}}
\put(5,5){\line(0,1){16}}
\put(5,21){\vector(1,-1){5}}
\put(43,13){\makebox(0,0){2}}
\put(43,13){\circle{5}}
\put(43,5){\makebox(0,0){<\p>}}
\put(37,13){\vector(-1,0){19}}
\put(28,28){\makebox(0,0){3}}
\put(28,28){\circle{5}}
\put(28,28){\circle{6}}
\put(36,28){\makebox(0,0){<\p>}}
\put(25,25){\vector(-1,-1){9}}
\put(31,25){\vector(1,-1){9}}
\end{picture}
\end{center}
Note that in world~<3>, the default atom~<\Not(\p)> is false, so
the interpretation of this world is non-minimal
(there is no need to make~<\p> true).  Thus, there
can be no incoming edges.

Let us now finally present the Kripke structure which yields
infinitely many models:
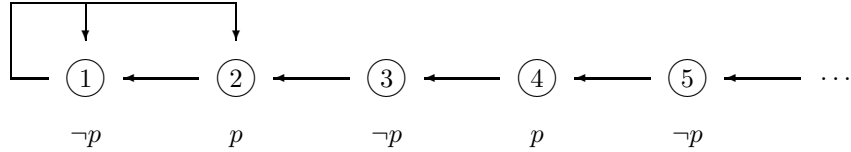
\begin{center}
\setlength{\unitlength}{1mm}
\begin{picture}(120,28)(0,0)
\put(15,13){\makebox(0,0){1}}
\put(15,13){\circle{5}}
\put(15,5){\makebox(0,0){<\lnot\p>}}
\put(10,13){\line(-1,0){5}}
\put(05,13){\line(0,1){10}}
\put(05,23){\line(1,0){30}}
\put(15,23){\vector(0,-1){5}}
\put(35,23){\vector(0,-1){5}}
\put(35,13){\makebox(0,0){2}}
\put(35,13){\circle{5}}
\put(35,5){\makebox(0,0){<\p>}}
\put(30,13){\vector(-1,0){10}}
\put(55,13){\makebox(0,0){3}}
\put(55,13){\circle{5}}
\put(55,5){\makebox(0,0){<\lnot\p>}}
\put(50,13){\vector(-1,0){10}}
\put(75,13){\makebox(0,0){4}}
\put(75,13){\circle{5}}
\put(75,5){\makebox(0,0){<\p>}}
\put(70,13){\vector(-1,0){10}}
\put(95,13){\makebox(0,0){5}}
\put(95,13){\circle{5}}
\put(95,5){\makebox(0,0){<\lnot\p>}}
\put(90,13){\vector(-1,0){10}}
\put(110,13){\vector(-1,0){10}}
\put(115,13){\makebox(0,0){\ldots}}
\end{picture}
\end{center}
Now we of course have to explain that different worlds really yield
different interpretations.  The trick is that world~<1> is the only
world in which neither~<\Not(\p)> nor~<\Not(\lnot\p)> are true.  Now other
worlds can be identified with the number of nested beliefs necessary
to get to world~1.  So the formula
\[{\mathcal B}^{\n-1}\parenB{\lnot\Not(\p)\land\lnot\Not(\lnot\p)}
\land{\mathcal B}^{\n-2} \parenB{\Not(\p)\lor\Not(\lnot\p)}\] is true in
world~<\n\geq2>, but in no other world.
Each world gives rise to one full model
(as explained in Theorem~\ref{thKripke} 
and Corollary~\ref{corKripke}),
so the theory <\set{\p\lif\Not(\p)}> really has infinitely many full models.
 
\end{example}

\section{Proof of the Model-theoretic Characterization
        (Theorem~\protect{\ref{TH:MODELTHEORETIC}})}
\label{app:modeltheoretic}
      
      We will first prove that a reduced
      model~<\redI=\objI\union\defI> with <\redI\models{\P}>
      and~<\defI\in\DefFix> can be extended to a full model~<\fullI> of
      the least static expansion~<\overline{\P}>.  Of course, this proof
      is based on the Kripke structure already mentioned in
      Section~\ref{sec:model}:

\begin{definition}[Standard Kripke Model]
\label{defStandardKripke}%
\mbox{}\\
Let <{\P}\subseteq{\mathcal L}^*_{\Not}> be a super program.  We call the Kripke
structure~<\K=\struct{\W,\R,\V}> defined as follows the ``standard
Kripke model'' of~<{\P}>.  Let <\DefFix> be the greatest fixpoint
of~<\DefToDef{\P}>.  Then:
\begin{enumerate}
\item The set of worlds~<\W> are the reduced
  interpretations~<\redI=\objI\union\defI> we are interested in,
  i.e.~satisfying <\redI\models{\P}> and~<\defI\in\DefFix>.
\item
<\R(\redI,\redI')>, i.e.~<\redI> sees~<\redI'> iff <\redI'> is a
minimal model of~<{\P}> and for all~<\p_1\until\p_\n\in\At>:
\[\redI\models\Not(\p_1\landUntil\p_\n)\metaThen
        \redI'\models\lnot\p_1\lorUntil\lnot\p_\n.\]
      \item The valuation~<\V(\redI)> of a reduced
        interpretation~<\redI=\objI\union\defI> is the objective
        part~<\objI>.
\end{enumerate}
\end{definition}

As before, we denote by~<\K(\redI)> the full interpretation satisfying
the default atoms true in world~<\redI>, i.e.{}
\[\K(\redI)\models\Not(\F)\iff
\mbox{for all~<\redI'\in\W> with~<R(\redI,\redI')>: }
(\K,\redI')\models\lnot\F.\] Now the reduced
interpretation~<\redI=\objI\union\defI> assigns a truth value to the
atoms <\Not(\p_1\landUntil\p_\n)> and the Kripke structure also
assigns a truth value to them.  But the construction guarantees that
the truth values always agree:

\begin{lemma}
\label{lemAgreeOnBeliefs}%
For <\redI=(\objI\union\defI)\in\W> and all~<\p_1\until\p_\n\in\At>:
\[\K(\redI)\models\Not(\p_1\landUntil\p_\n)
        \iff\defI\models\Not(\p_1\landUntil\p_\n).\]
\end{lemma}

\begin{proof}
  First we show the direction <\metaIf>: Let
  <\defI\models\Not(\p_1\landUntil\p_\n)>.  By the construction,
  <\redI'\models\lnot\p_1\lorUntil\lnot\p_\n> holds for all worlds~<\redI'> seen
  by~<\redI>, i.e.~satisfying <\R(\redI,\redI')>.  Thus,
  <\K(\redI)\models\Not(\p_1\landUntil\p_\n)>.
  
  Now we prove <\metaThen> by contraposition: Let
  <\defI\not\models\Not(\p_1\landUntil\p_\n)>.  Since <\DefFix> is a fixpoint
  of~<\DefToDef{\P}>, we have
  <\DefFix=\ObjToDef{\P}\parenB{\DefToObj{\P}(\DefFix)}>.  By the definition
  of~<\ObjToDef{\P}>, there is a
  non-empty~<\objSet'\subseteq\DefToObj{\P}(\DefFix)> such that the default
  atoms true in~<\defI> are the intersection of the corresponding true
  negations in~<\objSet'>.  Thus, there is <\objI'\in\objSet'> with
  <\objI'\models\p_1\landUntil\p_\n>.  Furthermore, for all
  <\q_1\until\q_\m> with <\defI\models\Not(\q_1\landUntil\q_\m)>, we have
  <\objI'\models\lnot\q_1\lorUntil\lnot\q_\m>.  Now by the definition
  of~<\DefToObj{\P}>, there is a~<\defI'\in\DefFix> such that
  <\redI'=\objI'\union\defI'> is a minimal model of~<{\P}>.  It follows
  that <\R(\redI,\redI')> holds, i.e.~<\redI> sees a world with
  valuation~<\objI'>, in which <\lnot\p_1\lorUntil\lnot\p_\n> is false.  Thus
  <\K(\redI)\not\models\Not(\p_1\landUntil\p_\n)>.
\end{proof}

\begin{lemma}
\label{lemExtensionIsModel}%
For all reduced interpretations~<\redI\in\W>, the full interpretation
<\fullI=\K(\redI)> is a model of the least static
expansion~<\overline{\P}>.
\end{lemma}

\begin{proof}
We show that the conditions of Theorem~\ref{thKripke} are satisfied:
\begin{enumerate}
\item We first have to show that every world~<\redI=\objI\union\defI>
  sees at least one world~<\redI'>.  This follows
  from~<\defI\in\DefFix>,
  i.e.~<\defI\in\ObjToDef{\P}\parenB{\DefToObj{\P}(\DefFix)}>.  By the
  definition of~<\ObjToDef{\P}> there is a non-empty
  subset~<\objSet\subseteq\DefToObj{\P}(\DefFix)> such that
  every~<\objI'\in\objSet> satisfies for all~<\p_1\until\p_\n\in\At>:
  \[\defI\models\Not(\p_1\landUntil\p_\n)\metaThen
  \objI'\models\lnot\p_1\lorUntil\lnot\p_\n.\]
  Now by the definition of~<\DefToObj{\P}>,
  there is a~<\defI'\in\DefFix> such that
  <\redI'=\objI'\union\defI'> is a minimal model of~<{\P}>.
  But then <\R(\redI,\redI')> holds.
\item The reduced interpretations~<\redI\in\W> satisfy the program~<{\P}>.
  By Lemma~\ref{lemAgreeOnBeliefs}, <\redI> and~<\fullI=\K(\redI)>
  agree on the atoms occurring in~<{\P}>.  Thus, also <\fullI> is a model
  of~<{\P}>.
\item
  For every~<(\redI,\redI')\in\R>,
  the reduced interpretation~<\redI'> is a minimal model of~<{\P}>.
  If the full interpretation~<\fullI'=\K(\redI')> were not a minimal model
  of~<{\P}>, i.e.~there were a smaller model~<\hat\fullI'> of~<{\P}>,
  then its reduct~<\hat\redI'> would be a smaller model than~<\redI'>
  (using again Lemma~\ref{lemAgreeOnBeliefs} in order to conclude that
  <\redI'> and~<\fullI'>, and thus <\redI'> and~<\hat\redI'> agree on
  the default atoms).
\end{enumerate}
Now Theorem~\ref{thKripke} allows us to conclude that the formulas
true in all worlds of~<\W> are a static expansion of~<{\P}>.  Of course,
every~<\fullI=\K(\redI)> is a model of this static expansion.  But the
least static expansion~<\overline{\P}> is a subset, so <\fullI> is also a
model of~<\overline{\P}>.
\end{proof}

\begin{lemma}
\label{lemCnNotImpl}%
<\Not(\lnot\F_1\landUntil\lnot\F_\m\land\G)\;\vdash_{\Not}\;
\Not(F_1)\landUntil\Not(F_\m)\lthen\Not(\G)>.
\end{lemma}

\begin{proof}
  This is a simple exercise in applying the axioms of~<{\mathcal
    L}_{\Not}>: First, we get
\[\Not\!\parenB{\F_1\lorUntil\F_\m\lor(\lnot\F_1\landUntil\lnot\F_\m\land\G)}
        \liff\Not(F_1\lorUntil\F_\m\lor\G)\]
by the invariance inference rule.
Then we apply on both sides the distributive axiom:
\[\begin{array}{@{}ll@{}}
        &\Not(\F_1)\landUntil\Not(\F_\m)\land
                \Not(\lnot\F_1\landUntil\lnot\F_\m\land\G)\\
        \liff&\Not(F_1)\landUntil\Not(\F_\m)\land\Not(\G).
\end{array}\]
This implies propositionally:
\[\Not(\F_1)\landUntil\Not(\F_\m)
        \land\Not(\lnot\F_1\landUntil\lnot\F_\m\land\G)
        \lthen\Not(\G).\]
Now we insert our precondition
and get the required formula.
\end{proof}

\begin{lemma}
\label{lemReductInFixpoint}%
Let <{\P}> be finite, <\fullI> be a full model of the least static
expansion~<\overline{\P}>, and let <\redI=\objI\union\defI> be its
reduct.  Then <\defI\in\DefFix>.
\end{lemma}

\begin{proof}
  We show by induction on~<\k> that the default part~<\defI> of a
  model~<\fullI> of the least static expansion~<\overline{\P}> is
  contained in~<\DefToDef{\P}^\k(\Def)>.  For <\k=0> this is trivial,
  since <\Def> is the complete set of default interpretations.
  
  Let us assume that <\defI\in\DefToDef{\P}^\k(\Def)>.  We have to show
  that~<\defI> is not ``filtered out'' by one further application
  of~<\DefToDef{\P}>.  Let <\defSet\defEq\DefToDef{\P}^\k(\Def)>,
  <\objSet\defEq\DefToObj{\P}(\defSet)>, and
\[\begin{array}{@{}l@{}l@{}}
        \objSet'\defEq\setCondB{\objI\in\objSet}{\:
        &\mbox{for every default atom~<\Not(\p_1\landUntil\p_\n)>:}
        \\
        &\mbox{\qquad if <\defI\models\Not(\p_1\landUntil\p_\n)>,}\\
        &\mbox{\qquad then <\objI\models\lnot\p_1\lorUntil\lnot\p_\n>}}.
\end{array}\]
We will now show that <\objSet'> has the properties required in the
definition of~<\ObjToDef{\P}>, namely we will show
\[\defI\models\Not(\p_1\landUntil\p_\n)
\iff\mbox{for all <\objI'\in\objSet'>: } \objI'\models\lnot\p_1\lorUntil\lnot\p_\n.\]
This implies that <\objSet'> is non-empty because for <\n=0> the empty
conjunction is logically true.  Because of the consistency axiom,
<\defI\not\models\Not(\true)>.  But then there must be at least
one~<\objI'\in\objSet'>, because otherwise the ``for all'' on the right
hand side would be trivially true.
Also the direction~<\metaThen> follows trivially from the construction.

Now we have to show that for every default
atom~<\Not(\p_1\landUntil\p_\n)> which is false in~<\defI>, there
is~<\objI'\in\objSet'> with~<\objI'\not\models\lnot\p_1\lorUntil\lnot\p_\n>.  Let
<\Not(\q_{{\i},1}\landUntil\q_{{\i},\n_{\i}})>, <{\i}=1\until\m>, be all
default atoms true in~<\defI> (containing only the finitely many
objective propositions occurring in~<{\P}>).  Since <\fullI\models\overline{\P}>,
the formula
\[\parenC{\landMulti{{\i}=1}{\m}
  \Not(\q_{{\i},1}\landUntil\q_{{\i},\n_{\i}})}
\;\lthen\;\Not(\p_1\landUntil\p_\n)\]
cannot be contained in~<\overline{\P}>
(it is violated by~<\defI> and thus by~<\fullI>).
But Lemma~\ref{lemCnNotImpl}
shows that the above formula would follow from
\[\Not\parenD{
  \parenC{\landMulti{{\i}=1}{\m} (\lnot\q_{{\i},1}\lorUntil\lnot\q_{{\i},\n_{\i}})}
  \land(\p_1\landUntil\p_\n)}.\] Thus, this formula is also not contained
in~<\overline{\P}>.  But the static semantics requires that
<\Not(\F)\in\overline{\P}> if <\overline{\P}\models_{\rm min}\lnot\F>.  So if
<\Not(\F)\not\in\overline{\P}>, there must be a minimal model
of~<\overline{\P}> violating <\lnot\F>, i.e.~satisfying~<\F>.  In our case
this means that there is a minimal model~<\fullI'> of~<\overline{\P}>
with <\fullI'\models\lnot\q_{{\i},1}\lorUntil\lnot\q_{{\i},\n_{\i}}> for~<{\i}=1\until\m>
and <\fullI'\models(\p_1\landUntil\p_\n)>,
i.e.~<\fullI'\not\models\lnot\p_1\lorUntil\lnot\p_\n>.  But by our inductive
hypothesis, the default part~<\defI'> of~<\fullI'> is contained
in~<\defSet>, and thus the objective part~<\objI'> is in~<\objSet>.
Since <\objI'\models\lnot\q_{{\i},1}\lorUntil\lnot\q_{{\i},\n_{\i}}>, it is contained
in~<\objSet'>.  Thus, <\objI'> is the required element.
\end{proof}

\vspace*{3mm}

Now Theorem~\ref{TH:MODELTHEORETIC} follows directly from
Lemma~\ref{lemExtensionIsModel} and Lemma~\ref{lemReductInFixpoint}:

\setcounter{savedSection}{\value{section}}
\setcounter{savedTheorem}{\value{theorem}}
\setcounter{section}{\value{savedSecModelTheoretic}}
\setcounter{theorem}{\value{savedThModelTheoretic}}
\def\thesection{\arabic{section}}

\begin{theorem}[Model-Theoretic Characterization]
\mbox{}\\
Let \mh{P} be a finite super program:
\begin{enumerate}
\item The operator~<\DefToDef{\P}> is monotone (in the lattice of subsets
  of~<\Def> with the order <\supseteq>) and thus its iteration beginning from
  the set~<\Def> (all interpretations of the default atoms, the bottom
  element of this lattice) has a fixed point~<\DefFix>.
\item <\DefFix> consists exactly of all <\defI\in\Def> that
  are default parts of a full model <\fullI> of <\overline{\P}>.
\item A reduced interpretation~<\objI\union\defI> is a reduct of a
  full model~<\fullI> of~<\overline{\P}> iff <\defI\in\DefFix> and
  <\objI\union\defI\models{\P}>.
\end{enumerate}
\end{theorem}

\setcounter{section}{\value{savedSection}}
\setcounter{theorem}{\value{savedTheorem}}
\def\thesection{\Alph{section}}

\begin{proof}
  The monotonicity of~<\DefToDef{\P}> is obvious: Let
  <\defSet_1\supseteq\defSet_2> (since we are working with inverse set
  inclusion, this means that <\defSet_1> is below~<\defSet_2> in the
  lattice).  Then we get
  <\DefToObj{\P}(\defSet_1)\supseteq\DefToObj{\P}(\defSet_2)>:
  any <\objI\in\DefToObj{\P}(\defSet_2)> is based on
  a interpretation <\defI\in\defSet_2>, and since
  <\defSet_2\subseteq\defSet_1>, we also have <\objI\in\DefToObj{\P}(\defSet_1)>.
  Now let <\objSet_1\defEq\DefToObj{\P}(\defSet_1)> and
  <\objSet_2\defEq\DefToObj{\P}(\defSet_2)>.  {}From
  <\objSet_1\supseteq\objSet_2> it also follows that
  <\DefToDef{\P}(\defSet_1)=\ObjToDef{\P}(\objSet_1)\supseteq
  \ObjToDef{\P}(\objSet_2)=\DefToDef{\P}(\defSet_2)>, since any
  <\objSet'\subseteq\objSet_2> used to construct
  <\defI\in\ObjToDef{\P}(\objSet_2)> is also a subset of~<\objSet_1>.
  
  Part~2 means that for every~<\defI\in\Def>:
\[\mbox{There is a full model~<\fullI> of~<\overline{\P}>
        with belief part~<\defI> }
        \iff\defI\in\DefFix.\]
The direction~<\metaThen> is Lemma~\ref{lemReductInFixpoint}, and the
direction~<\metaIf> follows from Lemma~\ref{lemExtensionIsModel} and
the fact that any default interpretation~<\defI> can be extended to a
reduced model~<\redI=\objI\union\defI> of~<{\P}> (by making all
objective atoms~<\At> true in~<\objI>).  But then <\redI\in\W>.

Part~3 requires that for every reduced
interpretation~<\redI=\objI\union\defI>:
\[\begin{array}{@{}l@{}}
\mbox{There is a full model~<\fullI> of~<\overline{\P}>
        with reduct~<\redI> }\\
        \hspace*{2cm}\iff
        \mbox{ <\defI\in\DefFix> and <\objI\union\defI\models{\P}>.}
\end{array}\]
Here the direction~<\metaIf> is Lemma~\ref{lemExtensionIsModel} and
the direction~<\metaThen> follows trivially from
Lemma~\ref{lemReductInFixpoint} and <{\P}\subseteq\overline{\P}>.
\end{proof}


\begin{remark}
  Note that we have used the finiteness of~<{\P}> only in the proof to
  Lemma~\ref{lemReductInFixpoint}.  So even in the infinite case, our
  model-theoretic construction yields models of the least static
  expansion, but it does not necessarily yield all reduced models.
  One example, where a difference might occur, is
  <{\P}=\setCond{\q_{\i}\lor\r_{\i}}{{\i}\in\natNum}
  \union\setCond{\r_{\i}\lor\p}{{\i}\in\natNum}>.  Our model-theoretic
  construction excludes an interpretation which makes all
  <\Not(\q_{\i})> true and <\Not(\p)> false.  It seems plausible that
  <\overline{\P}> allows such models, because it would need an ``infinite
  implication'' to exclude them.  But this question needs further
  research.
 
\end{remark}

\section{Proof of Fixpoint Characterization
        (Theorem~\protect{\ref{TH:COMPUTE}})}
\label{app:fixpoint}
      
      We prove Theorem~\ref{TH:COMPUTE} by using the model-theoretic
      characterization.  More specifically, we show that formulas
      <\Not(\E_1)\landUntil\Not(\E_\m)\lthen\Not(\E_0)> contained
      in~<{\P}^\n> characterize exactly the default interpretations
      remaining after~<\n> applications of~<\DefToDef{\P}>.

First we prove the following lemma which characterizes minimal models
of special knowledge bases.

\begin{lemma}
\label{lemTbel}%
Let <\T> be any knowledge base and <\T_\Not> be a set of formulae
which contain only default atoms.  Minimal models of~<\T\union\T_\Not>
are precisely those minimal models of\/~<\T> which satisfy~<\T_\Not>.
\end{lemma}

\begin{proof}
  Let <\fullI> be a minimal model of~<\T\union\T_\Not>.  Of course,
  <\fullI> is a model of~<\T> and of~<\T_\Not>.  Now suppose that
  there is a smaller model~<\fullI'> of~<\T>.  Since <\fullI>
  and~<\fullI'> do not differ in the interpretation of default atoms,
  <\fullI'> is also a model of~<\T_\Not>, and thus a model
  of~<\T\union\T_\Not>.  But this contradicts the assumed minimality
  of~<\fullI>.

Let now <\fullI> be a minimal model of~<\T> which also satisfies~<\T_\Not>.
Clearly, <\fullI> is a model of~<\T\union\T_\Not>.
Since <\fullI'> is also a model of~<\T>,
the existence of a smaller model~<\fullI'> would contradict the
minimality of~<\fullI>.
\end{proof}

Next, we need the monotonicity of the sequence~<{\P}^\n>:

\begin{lemma}
\label{lemComputeMonotonic}%
For every~<\n\in\natNum>: <{\P}^\n\subseteq{\P}^{\n+1}>.
\end{lemma}

\begin{proof}
  The propositional consequence operator~<\Cn> is monotonic and has no
  influence on the minimal models, so it suffices to show that the
  sequence <\hat{\P}_0\defEq{\P}>,
\[\begin{array}{@{}l@{}l@{}}
\hat{\P}^{\n+1}\defEq{\P}\union
        \setCond{&\Not\E_1\landUntil\Not\E_\m\lthen\Not\E_0}{\\
        &\hspace*{2cm}\hat{\P}^{\n}\models_{\min}
                \lnot\E_1\landUntil\lnot\E_\m\lthen\lnot\E_0}
\end{array}\]
increases monotonically.

The proof is by induction on~<\n>.  The case <\n=0> is trivial.  For
larger~<\n>, the inductive hypothesis gives us
<\hat{\P}^{\n}\supseteq\hat{\P}^{\n-1}> and all formulas in
<\hat{\P}^{\n}\setminus\hat{\P}^{\n-1}> contain only default atoms.  So
Lemma~\ref{lemTbel} yields that the minimal models of~<\hat{\P}^{\n}> are
a subset of the minimal models of~<\hat{\P}^{\n-1}>, and therefore
<\hat{\P}^{\n-1}\models_{\min}\lnot\E_1\landUntil\lnot\E_\m\lthen\E_0> implies
<\hat{\P}^\n\models_{\min}\lnot\E_1\landUntil\lnot\E_\m\lthen\E_0>,
i.e.~<\hat{\P}^{\n+1}\supseteq\hat{\P}^{\n}>.
\end{proof}

\vspace*{3mm}

Next, we have the small problem that we must in principle look at
infinitely many default negation atoms, although we have required our
program to be finite.  For instance, <\Not(\p)>, <\Not(\p\land\p)>, and
so on are different default atoms, and in general, default
interpretations could assign different truth values to them.  However,
already <{\P}^1> excludes this:

\begin{definition}[Regular Model]
  \mbox{}\\
  Let a super program~<{\P}> be given.  A default interpretation~<\defI>
  is called regular wrt~<{\P}> iff
\begin{enumerate}
\item
If <\defI\models\Not(\p_1\landUntil\p_\n)>
and <\set{\p_1\until\p_\n}\subseteq\set{\q_1\until\q_\m}>,
then <\defI\models\Not(\q_1\landUntil\q_\m)>.
\item
<\defI\models\Not(\p_1\landUntil\p_\n)>
if some <\p_{\i}\in\At> does not occur in~<{\P}>.
\end{enumerate}
\end{definition}

\begin{lemma}
\label{lemRegularModel}%
All <\defI\in\defSet_{\i}>, <{\i}\geq1> are regular.
\end{lemma}

\begin{proof}
\begin{enumerate}
\item Every interpretation~<\redI> satisfies
  <\lnot(\p_1\landUntil\p_\n)\lthen\lnot(\q_1\landUntil\q_\m)>, if
  <\set{\p_1\until\p_\n}\subseteq\set{\q_1\until\q_\m}>.  So
  <\Not(\p_1\landUntil\p_\n)\lthen\Not(\q_1\landUntil\q_\m)> is
  contained in all~<{\P}^{\i}>, <{\i}\geq1>.
\item Every minimal model~<\redI> of~<{\P}> satisfies
  <\lnot(\p_1\landUntil\p_\n)> if some~<\p_{\i}> does not occur in~<{\P}>
  (because <\redI\models\lnot\p_{\i}>).  But then
  <\Not(\p_1\landUntil\p_\n)\in{\P}^1>, and by
  Lemma~\ref{lemComputeMonotonic} it is contained also in
  every~<{\P}^\n>, <\n\geq1>.
\end{enumerate}
\end{proof}

\begin{lemma}
\label{lemComputeModTh}%
Let <P^\n_{\Not}> be the set of formulas of the form
\[\Not(\E_1)\landUntil\Not(\E_\m)\lthen\Not(\E_0)\]
contained in~<{\P}^\n>.  Furthermore, let <\defSet_0\defEq\Def> and
<\defSet_{\n+1}\defEq\DefToDef{\P}(\defSet_\n)>.  Then for every
<\redI=\objI\union\defI> with <\redI\models{\P}>:
<\defI\models{\P}^\n_{\Not}\iff\defI\in\defSet_\n>.
\end{lemma}

\begin{proof}
  The proof is by induction on~<\n>.  The case <\n=0> is trivial,
  since <\defSet_0=\Def> and <{\P}^0={\P}> and we anyway consider only
  models of~<{\P}>.
\begin{enumerate}
\item ``<\metaIf>'': Let
  <\defI\in\defSet_{\n+1}=\DefToDef{\P}(\defSet_\n)>.  We have to show
  that <\defI\models{\P}^{\n+1}_{\Not}>.  Suppose that this were not the case,
  i.e.~<\defI> violates a formula
  <\Not\E_1\landUntil\Not\E_\m\lthen\Not\E_0> contained
  in~<{\P}^{\n+1}_{\Not}>.  This means that <\defI\models\Not\E_{\i}>
  for~<{\i}=1\until\m>, but <\defI\not\models\Not\E_0>.  By the definition
  of~<\DefToDef{\P}>, there must be an objective
  model~<\objI'\in\DefToObj{\P}(\defSet_\n)> (contained in the
  non-empty~<\objSet'>) such that <\objI'\models\lnot\E_{\i}> for~<{\i}=1\until\m>
  and <\objI'\not\models\lnot\E_0>.  By the definition of~<\DefToObj{\P}>, there
  must be a default interpretation~<\defI'\in\defSet_\n> such that
  <\redI'=\objI'\union\defI'> is a minimal model of~<{\P}>.  Then the
  inductive hypothesis gives us <\defI'\models{\P}^\n_{\Not}>, and
  Lemma~\ref{lemTbel} allows us to conclude that <\redI'> is a minimal
  model of~<{\P}\union{\P}^\n_{\Not}> and thus of~<{\P}^\n>.  But this means
  that <{\P}^\n\not\models_{\min}\Not\E_1\landUntil\Not\E_\m\lthen\Not\E_0>.
  Since <\redI'> is a model of~<{\P}>, the critical formula cannot be
  contained in~<{\P}> (if <\E_0> is the empty conjunction and <{\P}> is not
  affirmative, this would be syntactically possible).  And finally, it
  cannot be introduced by the <\Cn>-operator, since <\redI'> is a
  model of its preconditions.  Thus, it is impossible that
  <\Not\E_1\landUntil\Not\E_\m\lthen\Not\E_0> is contained
  in~<{\P}^{\n+1}_{\Not}>.
\item ``<\metaThen>'': Let <\defI\models{\P}^{\n+1}_{\Not}>.  By
  Lemma~\ref{lemComputeMonotonic} we have
  <{\P}^{\n}_{\Not}\subseteq{\P}^{\n+1}_{\Not}>, so <\defI\models{\P}^{\n}_{\Not}>, and the
  inductive hypothesis gives us <\defI\in\defSet_\n>.  We have to show
  that <\defI> is not ``filtered out'' by one further application of
  the <\DefToDef{\P}>-operator.  Let
\[
\objSet'\defEq\setCondB{\objI'\in\DefToObj{\P}(\defSet_\n)}{
        \mbox{<\objI'\models\lnot\E>}
        \mbox{ for all <\E> with <\defI\models\Not\E>}}.
\]
We have to show that for every <\E_0> with <\defI\not\models\Not\E_0> there
is an~<\objI'\in\objSet'> with <\objI'\not\models\lnot\E_0>.  This especially
implies that <\objSet'> is non-empty, since <\defI\not\models\Not(\true)>,
which is obviously contained in~<{\P}^{\n+1}_{\Not}>.

Now suppose that this were not the case, i.e.~there were an~<\E_0>
such that there is no <\objI'\in\objSet'> with <\objI'\not\models\lnot\E_0>.

Let <\Not(\E_{\i})>, <{\i}=1\until\m>, be all belief atoms which are true
in~<\defI> and satisfy the following conditions (in order to make the
set finite): First, the conjunctions~<\E_{\i}> contain only
propositions~<\p\in\At> occurring in~<{\P}> (which was required to be
finite), and second, each~<\E_{\i}> contains each proposition at most
once.

We now prove that <\Not\E_1\landUntil\Not\E_\m\lthen\Not\E_0> is
contained in~<{\P}^{\n+1}_{\Not}>, which contradicts
<\defI\models{\P}^{\n+1}_{\Not}>.  Let <\redI'=\objI'\union\defI'> be any
minimal model of~<{\P}^\n>.  We have to show that it satisfies
<\lnot\E_1\landUntil\lnot\E_\m\lthen\lnot\E_0>.  The induction hypothesis gives us
<\defI'\in\defSet_\n>, and thus <\objI'\in\DefToObj{\P}(\defSet_\n)>.
Suppose that <\objI'\not\models\lnot\E_{\i}> for <{\i}=1\until\m>, since otherwise
the formula is trivially satisfied.  Since <\redI'> is a minimal model
and <\defI> is regular, this means that <\objI'\in\objSet'> (the
validity of~<\lnot\E_1\until\lnot\E_\m> implies the validity of all other
<\lnot\E> considered in the construction of~<\objSet'>).  But we have
assumed that no element of~<\objSet'> violates~<\lnot\E_0>,
thus~<\objI'\models\lnot\E_0>.
\end{enumerate}
\end{proof}

\begin{lemma}
\label{lemComputeSubset}%
<{\P}^{\n_0}\subseteq\overline{P}|{\mathcal L}^*_{\Not}>.
\end{lemma}

\begin{proof}
  We show <{\P}^\n_{\Not}\subseteq\overline{P}> by induction on~<\n> (this
  implies the above statement since <\overline{\P}> is closed under
  consequences and <{\P}^{\n_0}\subseteq{\mathcal L}^*_{\Not}>).
  
  For <\n=0> this is trivial since <{\P}\subseteq\overline{\P}>.  Now suppose that
  <{\P}^\n\models_{\min}\lnot\E_1\landUntil\lnot\E_\m\lthen\lnot\E_0>.  By Theorem~3.8
  in~\cite{BDP:ainote} and the definition of~<\CnNot>, <\overline{\P}>
  differs from <{\P}> only by the addition of formulas containing only
  default atoms plus propositional consequences.  By the inductive
  hypothesis, <{\P}^\n_{\Not}\subseteq\overline{P}>.  Now Lemma~\ref{lemTbel}
  gives us that all minimal models of~<\overline{\P}> are also minimal
  models of~<{\P}^\n>, and therefore
  <\overline{\P}\models_{\min}\lnot\E_1\landUntil\lnot\E_\m\lthen\lnot\E_0>,
  i.e.~<\overline{\P}\models_{\min}\lnot\E_1\landUntil\lnot\E_\m\land\E_0>.  This means
  that <\Not(\lnot\E_1\landUntil\lnot\E_\m\land\E_0)\in\overline{\P}>, and by
  Lemma~\ref{lemCnNotImpl} we get that
  <\Not\E_1\landUntil\Not\E_\m\lthen\Not\E_0> is contained
  in~<\overline{\P}>.
\end{proof}

\vspace*{3mm}

Now we can complete the proof of Theorem~\ref{TH:COMPUTE}.  First, a
fixpoint is reached after a finite number of iterations, because by
Lemma~\ref{lemRegularModel} we know that there are only a finite
number of ``really different'' default negation atoms, so after the
first iteration (which ensures the regularity) it suffices to consider
a finite number of implications
<\Not\E_1\landUntil\Not\E_\m\lthen\Not\E_0>, and the sets~<{\P}^\n> are
monotonically increasing (Lemma~\ref{lemComputeMonotonic}).

So we have <{\P}^{\n_0}_{\Not}={\P}^{\n_0+1}_{\Not}> and thus
<\defSet_{\n_0}=\defSet_{\n_0+1}=\DefFix>.  Now for any reduced
interpretation~<\redI=\objI\union\defI>: if <\redI\models{\P}^{\n_0}>, then
<\redI\models{\P}> and <\redI\models{\P}^{\n_0}_{\Not}>, and by
Lemma~\ref{lemComputeModTh} we get <\defI\in\DefFix>.  Now the already
proven Theorem~\ref{TH:MODELTHEORETIC} implies that <\redI> is a
reduct of a full model of the least static expansion~<\overline{\P}>.
But this implies <\redI\models\overline{P}|{\mathcal L}^*_{\Not}>.  The
other direction <\redI\models\overline{P}|{\mathcal
  L}^*_{\Not}\metaThen\redI\models{\P}^{\n_0}> follows from
Lemma~\ref{lemComputeSubset}.

{}From the equivalence of~<{\P}^{\n_0}> and~<\overline{P}|{\mathcal
  L}^*_{\Not}> we get <{\P}^{\n_0}=\overline{P}|{\mathcal L}^*_{\Not}>,
since both sets are closed under propositional consequences: For
instance, let <\F\in{\P}^{\n_0}> and suppose that
<\F\not\in\overline{P}|{\mathcal L}^*_{\Not}>.  Since
<\overline{P}|{\mathcal L}^*_{\Not}> is closed under propositional
consequences, there must be a reduced interpretation~<\redI> with
<\redI\models\overline{P}|{\mathcal L}^*_{\Not}>, but <\redI\not\models\F>.  This
is impossible since we already know that every model of
<\overline{P}|{\mathcal L}^*_{\Not}> is also a model of~<{\P}^{\n_0}>.

\begin{remark}
  The finiteness of~<{\P}> was used in the proof for
  <\defI\models{\P}^\n_{\Not}\metaThen\defI\in\defSet_\n>.  This was to be
  expected, since we strongly conjecture that for infinite programs,
  the model-theoretic construction does not yield all models of the
  static completion.  However, this does not give us any hint whether
  Theorem~\ref{TH:COMPUTE} might hold for infinite programs.  This
  question is topic of our future research.
 
\end{remark}

\section{Proofs of Properties Used in the Implementation}
\label{app:computation}

\setcounter{savedSection}{\value{section}}
\setcounter{savedTheorem}{\value{theorem}}
\setcounter{section}{\value{savedSecComputation}}
\setcounter{theorem}{\value{savedThHyperFix}}
\def\thesection{\arabic{section}}

\begin{theorem}\mbox{}\\
Let $P$ be a super logic program,
${\cal F}_0\defEq\emptyset$,
and ${\cal F}_{i+1}\defEq H_P({\cal F}_i)$,
and $n$ be a natural number such that ${\cal F}_{n+1}={\cal F}_n$.
Then the following holds for all Herbrand interpretations~$I$:
$I$ is a minimal model of the ground instantiation~$P^*$ of~$P$
if and only if $I$ is a minimal model of ${\cal F}_n$.
\end{theorem}

\setcounter{section}{\value{savedSection}}
\setcounter{theorem}{\value{savedTheorem}}
\def\thesection{\Alph{section}}

\begin{proof}
First we show that a (minimal) model of one of $P^*$ and ${\cal F}_n$
is also a model of the other:
\begin{enumerate}
\item
\label{case:hyperfixA}
The conditional facts in~$H_P({\cal F}_i)$
are logical consequences of~$P^*\union{\cal F}_i$.
By induction on~$i$,
it follows that ${\cal F}_i$ is implied by~$P^*$.
Therefore,
every model of~$P^*$ is also a model of~${\cal F}_n$.
\item
\label{case:hyperfixB}
Next,
we prove that every minimal model of~${\cal F}_n$
is also a model of~$P^*$:
Suppose that this would not be the case,
i.e.~$I$ is a minimal model of~${\cal F}_n$,
but it violates a rule
\[A_1\lor\ldots\lor A_k\leftarrow B_1\land\ldots\land B_m\land
		{\Not}C_1\land\ldots\land{\Not}C_l\]
in~$P^*$.
This means that $B_1\until B_m$ are true in~$I$.
Since $I$ is a minimal model of~${\cal F}_l$,
${\cal F}_l$ contains for $i=1\until m$ a conditional fact
${\cal A}_i\lif{\cal C}_i$
that is violated in the interpretation~$I\setminus\set{B_i}$
(i.e.~the interpretation that agrees with~$I$ except
that $B_i$ is false in it).
It follows that $B_i\in{\cal A}_i$,
that ${\cal A}_i\setminus\set{B_i}$ is false in~$I$,
and that ${\cal C}_i$ is true in~$I$
(since the default negation literals are treated like new propositions,
making $B_i$ false does not change any of the default negation literals).
Now consider the conditional fact
that is derived from the rule instance and the ${\cal A}_i\lif {\cal C}_i$:
\[\begin{array}{@{}l@{}}
\set{A_1\sigma,\ldots,A_k\sigma}\union
	({\cal A}_1\setminus\set{B_1\sigma})\union\ldots\union
	({\cal A}_m\setminus\set{B_m\sigma})
\;\lif\\
\hspace*{4cm}
\set{{\Not}C_1\sigma,\ldots,{\Not}C_l\sigma}
	\union{\cal C}_1\union\ldots\union{\cal C}_m
\end{array}\]
Since ${\cal F}_n$ is a fixpoint of~$H_P$,
this fact is also contained in~${\cal F}_n$.
But this is a contradiction,
since it is violated in~$I$.
\end{enumerate}
Now let $I$ be a minimal model of~$P^*$.
(\ref{case:hyperfixA}) shows that it is a model of~${\cal F}_n$.
If it were not minimal,
there would be a smaller model~$I_1$ of~${\cal F}_n$.
Then also a minimal model~$I_0$ of~${\cal F}_n$ must exist
that is still smaller than (or equal to)~$I_1$.
But by (\ref{case:hyperfixB}) above,
$I_0$ is also a model of~$P^*$
which contradicts the assumed minimality of~$I$.

Let conversely $I$ be a minimal model of~${\cal F}_n$.
By (\ref{case:hyperfixB}),
it is a model of~$P^*$.
If it were not minimal,
there would be a smaller model~$I_0$ of~$P^*$,
which is by~(\ref{case:hyperfixA}) also a model of~${\cal F}_n$,
which again contradicts the assumed minimality of~$I$.
\end{proof}

\begin{lemma}
Let $T_1$ be a nonmonotonic knowledge base with $T_1\models_{\min}F$.
Let $T_2$ be a set of default negation atoms,
i.e.~formulas of the form~$\Not G$ with an arbitrary formula~$G$.
Then $\CnNot(T_1\union T_2)\models_{\min} F$.
\end{lemma}

\begin{proof}
If this were not the case,
there were a minimal model~$I$ of~$\CnNot(T_1\union T_2)$
which does not satisfy~$\lneg F$.
Because of $T_1\models_{\min}\lneg F$ and
$T_1\subseteq \CnNot(T_1\union T_2)$,
there must be a model~$I_0$ of~$T_1$ that is smaller than~$I$,
but not a model of~$\CnNot(T_1\union T_2)$.
Since $I_0$ is smaller than~$I$,
it assignes the same truth values to the default negation literals.
But then it satisfies~$T_2$,
i.e.~it is a model of~$T_1\union T_2$.
Thus,
a violated formula must be one that is added by $\CnNot$.
Take the first such formula.
It cannot be a propositional consequence,
because propositional consequences are by definition satisfied
in all models that satisfy the preconditions
(and until this first formula, it satisfied all preconditions).
But it can also not be added by (CA), (DA), or (IR),
because all these formulas consist only of default negation literals,
which are interpreted the same in both models.
Therefore,
we can conclude
$I_0\models\CnNot(T_1\until T_2)$.
But that contradicts the assumed minimality of~$I$.
\end{proof}

\setcounter{savedSection}{\value{section}}
\setcounter{savedTheorem}{\value{theorem}}
\setcounter{section}{\value{savedSecComputation}}
\setcounter{theorem}{\value{savedThOptim}}
\def\thesection{\arabic{section}}

\begin{theorem}
\begin{enumerate}
\item
Let $T$ be a knowledge base
with $T\models_{\min}\lneg F$.
Then $T$ and $T\union\set{\Not F}$ have the same static expansions.
\item
Let $T_1$ and $T_2$ be knowledge bases with
$\CnNot(T_1)=\CnNot(T_2)$.
Then $T_1$ and $T_2$ have the same static expansions.
\end{enumerate}
\end{theorem}

\setcounter{section}{\value{savedSection}}
\setcounter{theorem}{\value{savedTheorem}}
\def\thesection{\Alph{section}}

\begin{proof}
\begin{enumerate}
\item
Let $T^\diamond$ be a static expansion of~$T$.
By the lemma above,
$T^\diamond\models_{\min}\lneg F$.
But then it easily follows that $T^\diamond$ is also a static expansion
of $T\union\set{\Not F}$:
It has to satisfy
\[T^\diamond=\CnNot\parenB{T\union\set{\Not F}\union
		\setCond{\Not G}{T^\diamond\models_{\min}\lneg G}}.\]
Since $T^\diamond\models_{\min} \lneg F$,
the formula $\Not F$ is anyway contained in the preconditions of~$\CnNot$,
so the union with $\set{\Not F}$ changes nothing.

Assume conversely that $T^\diamond$ is a static expansion
of $T\union\set{\Not F}$,
i.e.{}
\[T^\diamond=\CnNot\parenB{T\union\set{\Not F}\union
		\setCond{\Not G}{T^\diamond\models_{\min}\lneg G}}.\]
Again by the lemma above we get $T^\diamond\models_{\min} \lneg F$.
But this means that the preconditions are not changed
when we do not add $\set{\Not F}$ explicitly to the preconditions:
\[T^\diamond=\CnNot\parenB{T\union
		\setCond{\Not G}{T^\diamond\models_{\min}\lneg G}}.\]
\item
This follows with the following sequence of equations:
\[\begin{array}{@{}lcl@{}}
T^\diamond&=&\CnNot\parenB{T_1\union
		\setCond{\Not F}{T^\diamond\models_{\min}\lneg F}}\\
	&=&\CnNot\parenB{\CnNot(T_1)\union
		\setCond{\Not F}{T^\diamond\models_{\min}\lneg F}}\\
	&=&\CnNot\parenB{\CnNot(T_2)\union
		\setCond{\Not F}{T^\diamond\models_{\min}\lneg F}}\\
	&=&\CnNot\parenB{T_2\union
		\setCond{\Not F}{T^\diamond\models_{\min}\lneg F}}.
\end{array}\]
\end{enumerate}~
\end{proof}

\setcounter{savedSection}{\value{section}}
\setcounter{savedTheorem}{\value{theorem}}
\setcounter{section}{\value{savedSecComputation}}
\setcounter{theorem}{\value{savedThModGen}}
\def\thesection{\arabic{section}}

\begin{theorem}\mbox{}\\
Let ${\cal D}$ be a set of disjunctions of objective atoms,
which does not contain the empty disjunction~$\false$,
and which does not contain two disjunctions~${\cal A}$
and ${\cal A}'$,
such that ${\cal A}\subset{\cal A}'$ (i.e.~${\cal A}'$ is non-minimal).
Let $I$ be a partial interpretation
such that atoms interpreted as false do not appear in~${\cal D}$
and atoms interpreted as true appear as facts in~${\cal D}$.
Let $p$ be an atom that appears in
the proper disjunction
\[p_1\lorUntil p_{i-1}\lor p\lor p_{i+1}\lorUntil p_n\]
(and possibly more such disjunctions).
Then
\begin{enumerate}
\item
There is a minimal model of~${\cal D}$
that extends~$I$ and interprets $p$ as false.
\item
There is a minimal model of~${\cal D}$
that extends~$I$ and interprets~$p$ as true
and $p_1\until p_{i-1},p_{i+1}\until p_n$ as false.
\end{enumerate}
\end{theorem}

\setcounter{section}{\value{savedSection}}
\setcounter{theorem}{\value{savedTheorem}}
\def\thesection{\Alph{section}}

\begin{proof}
\begin{enumerate}
\item
Consider the interpretation~$I'$
that extends $I$ by interpreting $p$ as false and all remaining atoms as true.
Suppose that it were not a model of~${\cal D}$,
i.e.~it would violate some disjunction in~${\cal D}$.
Since all atoms that $I$ interprets as false do not appear in~${\cal D}$,
and all the remaining atoms except~$p$ are interpreted as true,
the violated disjunction can only be~$p$.
But this is a contradiction,
since then the proper disjunction
$p_1\lorUntil p_{i-1}\lor p\lor p_{i+1}\lorUntil p_n$
would not be minimal.
Thus,
$I'$ is a model of~${\cal D}$.
Then there is also a minimal model~$I_0$ of~${\cal D}$
that is less than or equal to~$I'$.
The atoms interpreted as false in~$I$ as well as $p$
must be false in~$I_0$, since it is less or equal to~$I'$.
The atoms interpreted as true in~$I$ appear as facts in~${\cal D}$,
so they must be true in~$I_0$.
\item
Consider the interpretation~$I'$
that extends $I$ by interpreting $p$ as true,
$p_1\until p_{i-1},p_{i+1}\until p_n$ as false,
and all remaining atoms as true.
Suppose that it were not a model of~${\cal D}$,
i.e.~it would violate some disjunction in~${\cal D}$.
Since all atoms that $I$ interprets as false do not appear in~${\cal D}$,
and all the remaining atoms except~$p_1\until p_{i-1},p_{i+1}\until p_n$
are interpreted as true,
the atoms in the violated disjunction can only be a subset of
$\set{p_1\until p_{i-1},p_{i+1}\until p_n}$.
But this contradicts the assumed minimality of the disjunction
$p_1\lorUntil p_{i-1}\lor p\lor p_{i+1}\lorUntil p_n$.
Thus,
$I'$ is a model of~${\cal D}$.
Then there is also a minimal model~$I_0$ of~${\cal D}$
that is less than or equal to~$I'$.
This means that the atoms interpreted as false
in~$I$ as well as $p_1\until p_{i-1},p_{i+1}\until p_n$
must be false in~$I_0$.
All atoms interpreted as true in~$I$
must be true in~$I_0$,
since they appear as facts in~${\cal D}$.
Finally,
also $p$ must be true in~$I_0$,
since otherwise it would violate the disjunction
$p_1\lorUntil p_{i-1}\lor p\lor p_{i+1}\lorUntil p_n$.
\end{enumerate}
\end{proof}

\end{document}